 \documentclass[11pt]{article}
 \usepackage{bbm}
 \usepackage{amsthm}
\usepackage{amsmath,amssymb}
\usepackage{amsfonts}            
\usepackage{mathrsfs}            
\usepackage[numbers,sort&compress]{natbib}
\usepackage{booktabs} 
\usepackage{bbm}
\usepackage{bm}
\usepackage{color,xcolor}
\usepackage{algorithm}  
\usepackage{algorithmicx}  
\usepackage{algpseudocode}
\usepackage{url}
\ifx\pdfoutput\undefined 
\usepackage{graphicx}
\DeclareGraphicsExtensions{.eps,.ps} \else
\usepackage{graphicx}
\DeclareGraphicsExtensions{.pdf} \fi
\graphicspath{{figure/}}
\usepackage{float}
\usepackage{subfigure}
\usepackage{diagbox}

\newtheorem{remark}{Remark}

\usepackage{amssymb}
\usepackage{epsfig}
\usepackage{times}
\title{Seamless Tracking of Group Targets and Ungrouped Targets Using Belief Propagation}

\begin{document}
\author{Xuqi Zhang, Fanqin Meng, Haiqi Liu, Xiaojing Shen and Yunmin Zhu}
\renewcommand{\thefootnote}{}
\footnotetext[1]{This work was supported in part by the NSFC No. 61673282 and the 2019RZJ04. \textit{(Corresponding author: Fanqin Meng.)}}
\footnotetext[2]{Xuqi Zhang, Haiqi Liu, Xiaojing Shen and Yunmin Zhu are with School of Mathematics, Sichuan University, Chengdu, Sichuan 610064, China (email: zxqcc@stu.scu.edu.cn, 411566818@qq.com, shenxj@scu.edu.cn, ymzhu@scu.edu.cn).}
\footnotetext[3]{Fanqin Meng is with Artificial Intelligence Key Laboratory of Sichuan Province, Yibin, Sichuan 644000, China (e-mail: mengfanqin2008@163.com).}

\renewcommand{\thefootnote}{\arabic{footnote}}
\date{}
 \maketitle
\begin{abstract}
	This paper considers the problem of tracking a large-scale number of group targets. Usually, multi-target in most tracking scenarios are assumed to have independent motion and are well-separated. However, for group target tracking (GTT), the targets within groups are closely spaced and move in a coordinated manner, the groups can split or merge, and the numbers of targets in groups may be large, which lead to more challenging data association, filtering and computation problems. Within the belief propagation (BP) framework, we propose a scalable group target belief propagation (GTBP) method by jointly inferring target existence variables, group structure, data association and target states. The method can efficiently calculate the approximations of the marginal posterior distributions of these variables by performing belief propagation on the devised factor graph. As a consequence, GTBP is capable of capturing the changes in group structure, e.g., group splitting and merging. Furthermore, we model the evolution of targets as the co-action of the group or single-target motions specified by the possible group structures and corresponding probabilities. This flexible modeling enables seamless and simultaneous tracking of multiple group targets and ungrouped targets. Particularly, GTBP has excellent scalability and low computational complexity. It not only maintains the same scalability as BP, i.e., scaling linearly in the number of sensor measurements and quadratically in the number of targets, but also only scales linearly in the number of preserved group partitions. Finally, numerical experiments are presented to demonstrate the effectiveness and scalability of the proposed GTBP method.
\end{abstract}

\noindent{\bf keywords:} 
Group target tracking, group structure, scalability, belief propagation, factor graph.

\section{Introduction}
Over the past forty years, many popular algorithms have been developed for the multi-target tracking (MTT) problem, e.g., joint probabilistic data association (JPDA) \cite{JPDA,BS-JPDA}, multiple hypothesis tracking (MHT) \cite{Reid,Mallick2013,Forty-MHT}, random finite set (RFS) based methods \cite{PHD,LMB}, where the targets are generally assumed to have independent motion. Recently, the GTT problem has aroused tremendous interest in many applications, e.g., aircraft formations \cite{UAV,poore2}, vehicle convoys \cite{overview1}, and groups of robots \cite{robot1}, etc. In these scenarios, the targets within groups are usually close spaced and have coordinated motion, the groups can split and merge, and there may be large numbers of individual targets within groups. Compare to MTT, tracking groups not only suffers the difficulties such as missed detections, clutters, and measurement origin uncertainty \cite{Mallick2013,zhu,DMHE}, but also encounters the group structure uncertainty caused by group merging or splitting \cite{Graph1}. Due to these limitations, directly using these popular MTT methods to track group targets may suffer severe data association ambiguity, frequent track crossings and high computational load.

Recent works on GTT mainly include  \cite{poore2,Graph1,CPHD,Gordon1,Godsill1,Lf,Leadership,Blackman1999,overview1}. Specifically, a multiple frame clustering tracking method within the MHT framework is proposed in \cite{poore2}, which uses clustering methods to partition the targets or measurements into groups and computes the measurement likelihoods by using cluster centroids. In \cite{Graph1}, the authors mainly introduce an evolving graph network to describe the group structure dynamics, and combine the sequential Monte Carlo method to tackle the GTT problem, where the data association is realized by JPDA. Within the RFS framework, a variant of the cardinalized probability hypothesis density filter \cite{CPHD} is proposed to deal with the GTT problem. Furthermore, some works on group state dynamics modeling are developed in \cite{Gordon1,Godsill1,Lf,Leadership}, and some other works on GTT can be seen in \cite{Blackman1999,overview1}. Additionally, there are some other researches that focus on a similar problem to GTT, namely the extended target tracking (ETT) problem \cite{RFS1,GP,RM2}, where the target may occupy multiple sensor resolution cells and thus can generate multiple measurements per time step. The two problems have certain similarities, but are different in some aspects. These studies on ETT primarily focus on estimating the kinematic states and the extent parameters of the targets of interest, while the tracking of the targets within groups is rarely involved.

Lately, a state-of-the-art BP method has drawn a lot of attention in the field of target tracking, which is also known as message passing or the sum-product algorithm \cite{FG-BP}. BP aims to compute the approximations of the marginal posterior probability density functions (pdfs) or probability mass functions (pmfs) for the variables of interest \cite{Max-Sum}. Due to the advantages of BP in estimation accuracy, computational complexity and implementation flexibility, it promotes the development of scalable target tracking algorithms \cite{BP-MTT1,BP-MTT2,BP-SLM,BP-Tuning,BP-MMTT1,BP-MMTT2,BP-Registration,BP-ETT1,BP-ETT2,BP-GTT}. On the whole, most of the studies on BP are developed from different perspectives in the context of MTT, e.g., scalable MTT with unknown number of targets \cite{BP-MTT1,BP-MTT2}, decentralized simultaneous cooperative self-localization and MTT \cite{BP-SLM}, maneuvering MTT \cite{BP-Tuning,BP-MMTT1,BP-MMTT2} and sensor registration \cite{BP-Registration}. Additionally, a scalable ETT algorithm is proposed in \cite{BP-ETT1}, which extends BP for tracking the targets that may generate multiple measurements. Later, a scalable detection and tracking algorithm for geometric ETT is developed in \cite{BP-ETT2}, which is able to jointly infer the geometric shapes of targets. For the GTT problem, a group expectation maximization belief propagation method is proposed to track a single coordinated group with a known number of targets \cite{BP-GTT}. This method is not suitable for tracking an unknown number of group targets, where groups may split and merge.

In this paper, we consider the GTT problem involving group splitting and merging, track initiation, data association and filtering. Our main contributions are summarized as follows:
\begin{itemize}
	\item We present a factor graph formulation for the GTT problem, and propose a scalable GTBP method by jointly inferring target existence variables, group structure, data association and target states. The group structure variable enables the description and capture of the group structure changes, e.g., group splitting and merging.
	
	\item The evolution of targets is modeled as the co-action of the group or single-target motions specified by possible group structures and corresponding probabilities. This flexible modeling makes it possible to track multiple group targets and ungrouped targets\footnote{To facilitate the distinction from grouped targets, we refer to multiple targets that have independent motion as ungrouped targets.} seamlessly and simultaneously. 
	
	\item GTBP has excellent scalability and low computational complexity that only scales linearly in the number of preserved group partitions, linearly in the number of sensor measurements, and quadratically in the number of targets.
\end{itemize}

Numerical results verify that GTBP not only has excellent scalability but also obtains better tracking performance in GTT. Thus, it is applicable for tracking a large number of group targets.

The rest of this paper is organized as follows. Section 2 briefly reviews the factor graph and BP, and then presents the problem formulation. Section 3 develops the GTBP method. Subsequently, a detailed particle-based implementation of GTBP is presented in Section 4. Numerical experiments and comparison results are given in Section 5. Lastly, we conclude this paper in Section 6.

\textit{Notation:} we use capital calligraphic letters and boldface lower-case characters to denote finite sets (e.g., $\mathcal{V}$) and vectors (e.g., $\mathbf{x}$), respectively. $\mathrm{I}(\cdot)$ denotes the indicator function that $\mathrm{I}(i)=1$ if $i=0$ and otherwise 0. For any set $\mathcal{V}$, $\mathcal{V}\backslash i$ is short for $\{i^{\prime}\in\mathcal{V}| i^{\prime}\neq i\}$, and $|\mathcal{V}|$ denotes the cardinality. Throughout this paper, we use $p(\cdot)$ and $p(\cdot|\cdot)$ as generic symbols for unconditional and conditional pdfs or pmfs or their mixtures. We denote $\int p(\mathbf{x})\mathrm{d}{(\mathbf{x} \backslash \mathbf{x}^{(i)})}$ as the summation or integration over $\mathbf{x}$ except $\mathbf{x}^{(i)}$ (i.e., for discrete or continuous random variables).
\section{Problem Formulation}
In this section, we briefly review factor graphs and the BP framework. Next, some basic assumptions are given and then we state the GTT problem to be solved.
\subsection{Factor Graphs and BP}
The factor graph is a graphical model to describe the factorization of pdfs \cite{FG-BP}. We denote $\mathcal{V}$ and $\mathcal{F}$ as the sets of the variable node $i$ and the factor node $\phi$ in a factor graph with respect to the random variable $\mathbf{x}^{(i)}$ and the factor $p_{\phi}$, respectively. In a factor graph, the variable node $i$ and the factor node $\phi$ are connected by an edge if and only if $\mathbf{x}^{(i)}$ is an argument of $p_{\phi}(\cdot)$. Let $\mathcal{F}_{i}$ and $\mathcal{V}_{\phi}$ denote the sets of the factor nodes connected with the variable node $i$ and the variable nodes connected with the factor node $\phi$, respectively. Consider that a posterior pdf $p(\mathbf{x}|\mathbf{z})$ can be factorized as \cite{FG-BP}
\begin{align*}
	p(\mathbf{x}|\mathbf{z})\propto\prod_{\phi \in \mathcal{F}} p_{\phi}\left(\mathbf{x}_{\phi}\right),
\end{align*}
where $\mathbf{x}$ and $\mathbf{x}_{\phi}$ are the stacked vectors of $\mathbf{x}^{(i)}$ for $i\in\mathcal{V}$ and $i\in\mathcal{V}_{\phi}$, respectively. According to the factorization, BP provides an efficient way of approximating the marginal distributions, which computes the message of each node in the factor graph and passes the node's message to the connected nodes \cite{FG-BP}. Specifically, if the variable node $i$ is connected with the factor node $\phi$, we denote $\varphi_{\phi \rightarrow i}(\mathbf{x}^{(i)})$ and $\upsilon_{i \rightarrow \phi}(\mathbf{x}^{(i)})$ as the message passed from the variable node $i$ to the factor node $\phi$ and the message passed from the factor node $\phi$ to the variable node $i$, respectively, which are given by
\begin{align}
	\label{ftv}
	\begin{split}
		\varphi_{\phi \rightarrow i}(\mathbf{x}^{(i)})=& \int p_{\phi}(\mathbf{x}_{\phi}) \prod_{i^{\prime} \in \mathcal{V}(\phi) \backslash i} \upsilon_{i^{\prime} \rightarrow \phi}(\mathbf{x}^{(i^{\prime})})\mathrm{d}{(\mathbf{x}_{\phi} \backslash \mathbf{x}^{(i)})},
		\\
		\upsilon_{i \rightarrow \phi}(\mathbf{x}^{(i)})=&\prod_{\phi^{\prime} \in \mathcal{F}(i) \backslash \phi} \varphi_{\phi^{\prime} \rightarrow i}(\mathbf{x}^{(i)}),
	\end{split}
\end{align}
where the symbol “$\rightarrow$” indicates the flow of the message. Eventually, for each variable node $i$, a belief $\widetilde{p}(\mathbf{x}^{(i)})$ is obtained by the product of all the incoming messages with the normalizing constraint such that $\int\widetilde{p}(\mathbf{x}^{(i)})\mathrm{d}{\mathbf{x}^{(i)}}=1$, which provides an approximation of the marginal posterior pdf $p(\mathbf{x}^{(i)}|\mathbf{z})$.
\subsection{System Model and Joint Posterior pdf}
\subsubsection{System Model}
At any time $k$, each potential target (PT)  is either a legacy PT (i.e., a PT survived from time $k-1$ to time $k$) or a new PT (i.e., a newly detected target at time $k$). That is, the PTs can be divided into two categories at each time instance, namely the legacy PTs and the new PTs. Let $\underline{\mathbf{x}}_{k}^{(i)}$ be the state vector of the legacy PT $i$ at time $k$, consisting of the target position and possibly further parameters (e.g., velocity and acceleration), where $i\in\{1,\ldots,n_{k}\}$ and $n_{k}$ is the number of the legacy PTs at time $k$. The detection of the legacy PTs are modeled by the binary existence variables $\underline{r}_{k}^{(i)}\in\{0,1\}$, i.e., legacy PT $i$ exists at time $k$ if and only if $\underline{r}_{k}^{(i)}=1$. We denote $\underline{\mathbf{x}}_{k}$ and $\underline{\mathbf{r}}_{k}$ as the joint state vector and existence vector of the legacy PTs at time $k$, respectively, 
\begin{align*}
	\underline{\mathbf{x}}_{k}&:=\left[\underline{\mathbf{x}}_{k}^{(1)\mathrm{T}},\ldots,\underline{\mathbf{x}}_{k}^{(n_{k})\mathrm{T}}\right]^{\mathrm{T}},
	\\
	\underline{\mathbf{r}}_{k}&:=\left[\underline{r}_{k}^{(1)},\ldots,\underline{r}_{k}^{(n_{k})}\right]^{\mathrm{T}}.
\end{align*}
Assume that at time $k$, the sensor generates $m_{k}$ measurements and the joint measurement vector is
\begin{align*}
	\mathbf{z}_{k}&:=\left[\mathbf{z}_{k}^{(1)\mathrm{T}},\ldots,\mathbf{z}_{k}^{(m_{k})\mathrm{T}}\right]^{\mathrm{T}},
\end{align*}
where each measurement $\mathbf{z}_k^{(m)}$ either originates from a PT or random clutter. Basically, it is assumed that at any time $k$, a target can generate at most one measurement, and a measurement originates from at most one target. To incorporate the newly detected targets at time $k$, $m_{k}$ new PT states $\overline{\mathbf{x}}_{k}^{(m)}$, $m=1,\ldots,m_{k}$ are introduced, where each $\overline{\mathbf{x}}_{k}^{(m)}$ corresponds to the measurement $\mathbf{z}_{k}^{(m)}$. The detection of the new PTs are also modeled by the binary existence variables $\overline{r}_{k}^{(m)}\in\{0,1\}$, i.e., a measurement $\mathbf{z}_{k}^{(m)}$ is generated by a new PT $\overline{\mathbf{x}}_{k}^{(m)}$ if and only if $\overline{r}_{k}^{(m)}=1$. We denote $\overline{\mathbf{x}}_{k}$ and $\overline{\mathbf{r}}_{k}$ as the joint state vector and existence vector of the new PTs, respectively,
\begin{align*}
	\overline{\mathbf{x}}_{k}&:=\left[\overline{\mathbf{x}}_{k}^{(1)\mathrm{T}},\ldots,\overline{\mathbf{x}}_{k}^{(m_{k})\mathrm{T}}\right]^{\mathrm{T}},
	\\
	\overline{\mathbf{r}}_{k}&:=\left[\overline{r}_{k}^{(1)},\ldots,\overline{r}_{k}^{(m_{k})}\right]^{\mathrm{T}}.
\end{align*}
Notably, the new PTs at time $k$ become the legacy PTs at time $k+1$ when receiving new measurements, which means that the number of the legacy PTs at time $k+1$ is updated by $n_{k+1}=n_{k}+m_{k}.$ Since the number of PTs would increase with the accumulation of sensor measurements, we consider at most $N_{\text{max}}$ PTs at any time and perform a pruning step at each time step to remove unlikely PTs. That is, $N_{\text{max}}$ is the maximum possible number of PTs and the number of actual targets is not larger than $N_{\text{max}}$.

\begin{figure}[hbtp]
	\centering 
	\includegraphics[width=0.6\linewidth]{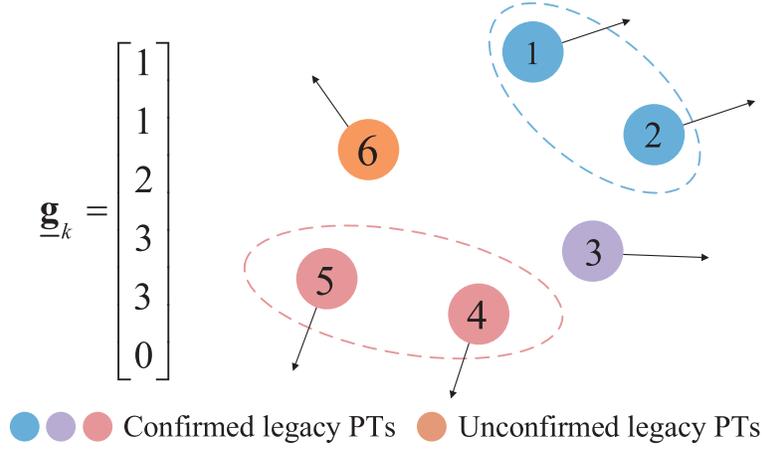}
	\caption{An example of $\underline{\mathbf{g}}_{k}$ for 5 confirmed legacy PTs (partitioned into two groups and an ungrouped target) and 1 unconfirmed legacy PT.}
	\label{figure1}
\end{figure}

In GTT, the group structure describes the connection between targets, which is a premise of the modeling of the evolution of targets. In this paper, we make the convention that at any time $k$, only the group structure of the confirmed legacy PTs (that have been declared to exist at the current time) is considered, and each confirmed legacy PT can be partitioned to only one group in a possible group structure. Concretely, we use a group partition vector $\underline{\mathbf{g}}_{k}:=\left[\underline{g}_{k}^{(1)},\ldots,\underline{g}_{k}^{(n_{k})}\right]^{\mathrm{T}}$ to represent the group structure of all legacy PTs at time $k$, and let $N(\underline{\mathbf{g}}_{k}):=\max(\underline{\mathbf{g}}_{k})$ denote the number of groups partitioned by $\underline{\mathbf{g}}_{k}$. For instance, $\underline{\mathbf{g}}_{k}:=\left[1\ 1\ 2\ 3\ 3\ 0\right]^{\mathrm{T}}$ (see Fig. \ref{figure1}) represents that the confirmed legacy PTs $\underline{\mathbf{x}}_{k}^{(1)}$ and $\underline{\mathbf{x}}_{k}^{(2)}$ are partitioned into group 1, $\underline{\mathbf{x}}_{k}^{(3)}$ is an ungrouped target (i.e., a single target), $\underline{\mathbf{x}}_{k}^{(4)}$ and $\underline{\mathbf{x}}_{k}^{(5)}$ are partitioned into group 3, and $\underline{\mathbf{x}}_{k}^{(6)}$ is an unconfirmed legacy PT. Furthermore, the group structure of the new PTs at time $k$ is represented by a variable $\overline{\mathbf{g}}_{k}$ with zero entries.

The unknown association between legacy PTs and measurements at time $k$ can be described by a target-oriented association vector $\mathbf{a}_{k}:=\left[a_{k}^{(1)},\ldots,a_{k}^{(n_{k})}\right]^{\mathrm{T}}$ with
\begin{align*}
	a_{k}^{(i)}:=
	\begin{cases}
		m \in\left\{1, \ldots, m_{k}\right\}, &\begin{array}{l}\text { if at time } k, \text { the PT } \underline{\mathbf{x}}_k^{(i)}  \text { generates the measurement } \mathbf{z}_k^{(m)}
		\end{array} \\
		0, &\begin{array}{l}\text { if at time } k, \text { the PT } \underline{\mathbf{x}}_k^{(i)} 
			\text { does not }
			\text{ generate a }
			\text{ measurement. }\end{array}
	\end{cases}
\end{align*}

\subsubsection{Joint Posterior pdf}
We denote the joint vectors of all the PT state, the existence variable and the group partition at time $k$ as $\mathbf{x}_k:=\left[\underline{\mathbf{x}}_{k}^{\mathrm{T}}, \overline{\mathbf{x}}_{k}^{\mathrm{T}}\right]^{\mathrm{T}}$,  $\mathbf{r}_k:=\left[\underline{\mathbf{r}}_{k}^{\mathrm{T}}, \overline{\mathbf{r}}_{k}^{\mathrm{T}}\right]^{\mathrm{T}}$ and $\mathbf{g}_k:=\left[\underline{\mathbf{g}}_{k}^{\mathrm{T}}, \overline{\mathbf{g}}_{k}^{\mathrm{T}}\right]^{\mathrm{T}}$, respectively. Let $\mathcal{R}_{k}$,  $\underline{\mathcal{R}}_{k}$ and $\underline{\mathcal{G}}_{k}$ be the sets of all possible $\mathbf{r}_{k}$, $\underline{\mathbf{r}}_{k}$ and $\underline{\mathbf{g}}_{k}$, respectively. For notational convenience, we define the augmented state vectors for the legacy PTs and the new PTs as
\begin{align*}
	\underline{\mathbf{y}}_{k}^{(i)}&:=\left[\underline{\mathbf{x}}_{k}^{(i)\mathrm{T}},\underline{r}_{k}^{(i)}\right]^{\mathrm{T}},
	\\
	\overline{\mathbf{y}}_{k}^{(m)}&:=\left[\overline{\mathbf{x}}_{k}^{(m)\mathrm{T}},\overline{r}_{k}^{(m)}\right]^{\mathrm{T}},
\end{align*}
and the joint augmented state vector at time $k$ is given by  $\mathbf{y}_{k}:=\left[\underline{\mathbf{y}}_{k}^{\mathrm{T}}, \overline{\mathbf{y}}_{k}^{\mathrm{T}}\right]^{\mathrm{T}}$. 

Let $\mathbf{y}_{1:k}$, $\mathbf{g}_{1:k}$, $\mathbf{a}_{1:k}$, $\mathbf{z}_{1:k}$ and $\mathbf{m}_{1:k}$ denote the stacked vectors of joint augmented states, group partition vectors, target-oriented association vectors, measurements and numbers of measurements up to time $k$, respectively. Assume that given all the PT states $\mathbf{x}_{k}$, the measurements $\mathbf{z}_{k}$
are conditionally independent of all
the past and future measurements $\mathbf{z}_{k^{\prime}}$ and PT states $\mathbf{x}_{k^{\prime}}$, $k^{\prime}\neq k$. By the chain rule and the conditional independence assumption, the posterior pdf $p(\mathbf{y}_{1:k}, \mathbf{g}_{1:k}, \mathbf{a}_{1:k}|\mathbf{z}_{1:k})$ can be obtained by
\begin{align}\label{pdf}
	\begin{split}
		p(\mathbf{y}_{1:k}, \mathbf{g}_{1:k}, \mathbf{a}_{1:k}|\mathbf{z}_{1:k})
		&=p(\mathbf{y}_{1:k}, \mathbf{g}_{1:k}, \mathbf{a}_{1:k}|\mathbf{z}_{1:k},\mathbf{m}_{1:k}) 
		\\
		&\propto p(\mathbf{z}_{1:k},  \mathbf{a}_{1:k},\mathbf{m}_{1:k}, \mathbf{y}_{1:k}, \mathbf{g}_{1:k}) 
		\\
		&=\prod_{k^{\prime}=1}^{k}p(\mathbf{z}_{k^{\prime}}, \mathbf{a}_{k^{\prime}},m_{k^{\prime}}, \mathbf{y}_{k^{\prime}}, \mathbf{g}_{k^{\prime}}|\mathbf{y}_{k^{\prime}-1}, \mathbf{g}_{k^{\prime}-1}), 
	\end{split}
\end{align}
where
\begin{align}\label{condition-pdf}
	\begin{split}
		p(\mathbf{z}_{k}, \mathbf{a}_{k},m_{k}, \mathbf{y}_{k},\mathbf{g}_{k}|\mathbf{y}_{k-1},\mathbf{g}_{k-1})=p(\mathbf{z}_{k}, \mathbf{a}_{k},m_{k}, \overline{\mathbf{y}}_{k},\overline{\mathbf{g}}_{k} |\underline{\mathbf{y}}_{k},\underline{\mathbf{g}}_{k}) p( \underline{\mathbf{y}}_{k},\underline{\mathbf{g}}_{k}|\mathbf{y}_{k-1},\mathbf{g}_{k-1}).
	\end{split}
\end{align}
Subsequently, we derive this joint posterior pdf under some regular assumptions. 
\subsection{Augmented State and Group Structure Transition pdf}
The augmented state and group structure transition pdf $p( \underline{\mathbf{y}}_{k},\underline{\mathbf{g}}_{k}|\mathbf{y}_{k-1},\mathbf{g}_{k-1})$ in (\ref{condition-pdf}) can be written as
\begin{align}\label{ps}
	\begin{split}
		p( \underline{\mathbf{y}}_{k},\underline{\mathbf{g}}_{k}|\mathbf{y}_{k-1},\mathbf{g}_{k-1})=p( \underline{\mathbf{y}}_{k}|\underline{\mathbf{g}}_{k}, \mathbf{y}_{k-1},\mathbf{g}_{k-1})p(\underline{\mathbf{g}}_{k}|\mathbf{y}_{k-1},\mathbf{g}_{k-1}).
	\end{split}
\end{align}
We assume that there is no PT at time $k=0$, i.e., $\mathbf{y}_{0}$, $\underline{\mathbf{y}}_{1}$ and $\underline{\mathbf{g}}_{1}$ are empty. For future use, we make convention that $p( \underline{\mathbf{y}}_{1}|\underline{\mathbf{g}}_{1}, \mathbf{y}_{0}):=1$ and $p(\underline{\mathbf{g}}_{1}|\mathbf{y}_{0}):=1$. Let $\Lambda_{\underline{\mathbf{g}}_{k}}(j)$ denote the index set of the PTs belonging to the group $j$ in the group partition $\underline{\mathbf{g}}_{k}$, i.e.,
\begin{align*}
	\Lambda_{\underline{\mathbf{g}}_{k}}(j):=\{i: \underline{\mathbf{g}}_{k}^{(i)}=j, i=1,\ldots,n_{k}\}.
\end{align*}
Note that $\Lambda_{\underline{\mathbf{g}}_{k}}(0)$ denotes the index set of the unconfirmed legacy PTs at time $k$. We use $\underline{\mathbf{y}}_{k,\Lambda_{\underline{\mathbf{g}}_{k}}(j)}$ to represent the joint augmented state of the PTs $i\in\Lambda_{\underline{\mathbf{g}}_{k}}(j)$. Assume that each group or unconfirmed legacy PT evolves independently of the other groups and unconfirmed legacy PTs, we have
\begin{align}\label{evolve}
	\begin{split}
		p(\underline{\mathbf{y}}_{k}|\underline{\mathbf{g}}_{k}, \mathbf{y}_{k-1},\mathbf{g}_{k-1})=
		\prod_{j=0}^{N(\underline{\mathbf{g}}_{k})}p(\underline{\mathbf{y}}_{k,\Lambda_{\underline{\mathbf{g}}_{k}}(j)}|\mathbf{y}_{k-1,\Lambda_{\underline{\mathbf{g}}_{k}}(j)}),
	\end{split}
\end{align}
where the augmented state transition density of the unconfirmed legacy PTs $p(\underline{\mathbf{y}}_{k,\Lambda_{\underline{\mathbf{g}}_{k}}(0)}|\mathbf{y}_{k-1,\Lambda_{\underline{\mathbf{g}}_{k}}(0)})$ is given by
\begin{align*}
	p(\underline{\mathbf{y}}_{k,\Lambda_{\underline{\mathbf{g}}_{k}}(0)}|\mathbf{y}_{k-1,\Lambda_{\underline{\mathbf{g}}_{k}}(0)})=\prod_{i\in\Lambda_{\underline{\mathbf{g}}_{k}}(0)}p(\underline{\mathbf{y}}_{k}^{(i)}|\mathbf{y}_{k-1}^{(i)}),
\end{align*}
with $p(\underline{\mathbf{y}}_{k}^{(i)}|\mathbf{y}_{k-1}^{(i)})=p(\underline{\mathbf{x}}_{k}^{(i)},\underline{r}_{k}^{(i)}|\mathbf{x}_{k-1}^{(i)}, r_{k-1}^{(i)})$ and
\begin{align}\label{sg}
	\begin{split}
		p(\underline{\mathbf{x}}_{k}^{(i)},\underline{r}_{k}^{(i)}|\mathbf{x}_{k-1}^{(i)}, r_{k-1}^{(i)}):=
		\begin{cases}
			f_{\mathrm{d}}(\underline{\mathbf{x}}_{k}^{(i)}), &\begin{array}{l} r_{k-1}^{(i)}=0,\  \underline{r}_{k}^{(i)}=0
			\end{array} 
			\\
			0, &\begin{array}{l}r_{k-1}^{(i)}=0,\  \underline{r}_{k}^{(i)}=1\end{array}
			\\
			(1-p_{\mathrm{s}}(\mathbf{x}_{k-1}^{(i)}))f_{\mathrm{d}}(\underline{\mathbf{x}}_{k}^{(i)}), &\begin{array}{l}r_{k-1}^{(i)}=1,\  \underline{r}_{k}^{(i)}=0\end{array}
			\\
			p_{\mathrm{s}}(\mathbf{x}_{k-1}^{(i)})p(\underline{\mathbf{x}}_{k}^{(i)}|\mathbf{x}_{k-1}^{(i)}), &\begin{array}{l}r_{k-1}^{(i)}=1,\  \underline{r}_{k}^{(i)}=1,\end{array}
		\end{cases}
	\end{split}
\end{align}
where $f_{\mathrm{d}}(\underline{\mathbf{x}}_{k}^{(i)})$ denotes a dummy pdf, $p(\underline{\mathbf{x}}_{k}^{(i)}|\mathbf{x}_{k-1}^{(i)})$ is the single-target state transition density, and $p_{\mathrm{s}}(\mathbf{x}_{k-1}^{(i)})$ is the survival probability of the PT $\mathbf{x}_{k-1}^{(i)}$ with $r_{k-1}^{(i)}=1$ and $\underline{r}_{k}^{(i)}=1$. Note that if there is a PT $\mathbf{x}_{k-1}^{(i)}, i\in\Lambda_{\underline{\mathbf{g}}_{k}}(j)$ with $r_{k-1}^{(i)}=0$, then it cannot exist at time $k$, i.e., $\underline{r}_{k}^{(i)}=0$. 

Furthermore, the pdf $p(\underline{\mathbf{y}}_{k,\Lambda_{\underline{\mathbf{g}}_{k}}(j)}|\mathbf{y}_{k-1,\Lambda_{\underline{\mathbf{g}}_{k}}(j)})$, $j\neq0$ describing the augmented state transition density of the group $j$ in the group partition $\underline{\mathbf{g}}_{k}$ can be factorized as
\begin{align}\label{GT-pdf2}
	\begin{split}
		p(\underline{\mathbf{y}}_{k,\Lambda_{\underline{\mathbf{g}}_{k}}(j)}|\mathbf{y}_{k-1,\Lambda_{\underline{\mathbf{g}}_{k}}(j)})=\big(\prod_{i\in\Lambda_{\underline{\mathbf{g}}_{k}}(j)\backslash \tilde{\Lambda}_{\underline{\mathbf{g}}_{k}}(j)}p(\underline{\mathbf{y}}_{k}^{(i)}|\mathbf{y}_{k-1}^{(i)})\big)\big(\prod_{i\in \tilde{\Lambda}_{\underline{\mathbf{g}}_{k}}(j)}p_{\mathrm{s}}(\mathbf{x}_{k-1}^{(i)})\big)p(\underline{\mathbf{x}}_{k,\tilde{\Lambda}_{\underline{\mathbf{g}}_{k}}(j)}|\mathbf{x}_{k-1,\tilde{\Lambda}_{\underline{\mathbf{g}}_{k}}(j)}),
	\end{split}
\end{align}
where $\tilde{\Lambda}_{\underline{\mathbf{g}}_{k}}(j):=\{i: r_{k-1}^{(i)}=\underline{r}_k^{(i)}=1, i\in\Lambda_{\underline{\mathbf{g}}_{k}}(j)\}$ is the index set of survival PTs in the group $j$, and  $\underline{\mathbf{x}}_{k,\Lambda_{\underline{\mathbf{g}}_{k}}(j)}$ denotes the joint state of the group $j$. The group transition density $p(\underline{\mathbf{x}}_{k,\tilde{\Lambda}_{\underline{\mathbf{g}}_{k}}(j)}|\mathbf{x}_{k-1,\tilde{\Lambda}_{\underline{\mathbf{g}}_{k}}(j)})$ describes the evolution of the group $j$, which degrades to the single-target state transition density if there is only one PT in the group.
\begin{remark}
	The group structure can be viewed as an attribute implicit of the targets in GTT, which determines the partition of targets into group targets and ungrouped targets. In this paper, we model the state transition by using group or single-target motion models according to the given group structure (\ref{evolve})-(\ref{GT-pdf2}), which enables seamlessly and simultaneously tracking of multiple group targets and ungrouped targets.
\end{remark}
Since the modeling of group dynamics is not the focus of this paper, here we apply the virtual leader-follower model \cite{Gordon1,RFS1} to describe the evolution of group targets, and some other models can refer to \cite{Godsill1,Lf,Leadership}. The model \cite{Gordon1,RFS1} assumes that the deterministic state of any target is a translational offset of the average state (i.e., the virtual leader) of the group. More specifically, let $\triangle\mathbf{x}_{k-1}^{(i)}$ denote the offset from the PT $\mathbf{x}_{k-1}^{(i)}\in\mathbf{x}_{k-1,\tilde{\Lambda}_{\underline{\mathbf{g}}_{k}}(j)}$ to the
virtual leader of the group $j$, i.e.,
\begin{align}\label{offset}
	\triangle\mathbf{x}_{k-1}^{(i)}:=\mathbf{x}_{k-1}^{(i)}-\breve{\mathbf{x}}_{k-1,\tilde{\Lambda}_{\underline{\mathbf{g}}_{k}}(j)},
\end{align}
where $\breve{\mathbf{x}}_{k-1,\tilde{\Lambda}_{\underline{\mathbf{g}}_{k}}(j)}$ is the virtual leader of the group $j$, i.e.,
\begin{align}\label{vl}
	\breve{\mathbf{x}}_{k-1,\tilde{\Lambda}_{\underline{\mathbf{g}}_{k}}(j)}&:=\frac{1}{|\tilde{\Lambda}_{\underline{\mathbf{g}}_{k}}(j)|}\sum_{i\in\tilde{\Lambda}_{\underline{\mathbf{g}}_{k}}(j)}\mathbf{x}_{k-1}^{(i)},
\end{align}
then the state transition model for the PT $i$ is
\begin{align}\label{model}
	\underline{\mathbf{x}}_{k}^{(i)}=f_{\mathrm{t}}(\breve{\mathbf{x}}_{k-1,\tilde{\Lambda}_{\underline{\mathbf{g}}_{k}}(j)})+\triangle\mathbf{x}_{k-1}^{(i)}+\mathbf{v}_{k}^{(i)},
\end{align}
where $f_{\mathrm{t}}(\cdot)$ is the state transition function of the virtual leader, and $\mathbf{v}_{k}^{(i)}$ are independent and identically distributed random variables with known pdf. Thus, the group transition density in (\ref{GT-pdf2}) can be written as
\begin{align}\label{pdf-model}
	\begin{split}
		p(\underline{\mathbf{x}}_{k,\tilde{\Lambda}_{\underline{\mathbf{g}}_{k}}(j)}|\mathbf{x}_{k-1,\tilde{\Lambda}_{\underline{\mathbf{g}}_{k}}(j)})=\prod_{i\in\tilde{\Lambda}_{\underline{\mathbf{g}}_{k}}(j)}p(\underline{\mathbf{x}}_{k}^{(i)}|\breve{\mathbf{x}}_{k-1,\tilde{\Lambda}_{\underline{\mathbf{g}}_{k}}(j)},\triangle\mathbf{x}_{k-1}^{(i)}),
	\end{split}
\end{align}
where $p(\underline{\mathbf{x}}_{k}^{(i)}|\breve{\mathbf{x}}_{k-1,\tilde{\Lambda}_{\underline{\mathbf{g}}_{k}}(j)},\triangle\mathbf{x}_{k-1}^{(i)})$ is described by the system model (\ref{model}).

The group structure transition pmf $p(\underline{\mathbf{g}}_{k}|\mathbf{y}_{k-1},\mathbf{g}_{k-1})$ in (\ref{ps}) determines how the information from $\mathbf{y}_{k-1}$ and $\mathbf{g}_{k-1}$ at time $k-1$ are used to guide the group structure changes. In this paper, we adopt a state-dependent model $p(\underline{\mathbf{g}}_{k}|\mathbf{y}_{k-1},\mathbf{g}_{k-1}):=p(\underline{\mathbf{g}}_{k}|\mathbf{x}_{k-1},\mathbf{r}_{k-1})$ for the group structure, i.e., $\underline{\mathbf{g}}_{k}$ is independent of $\underline{\mathbf{g}}_{k-1}$ given $\mathbf{y}_{k-1}$. Some similar models can refer to \cite{Godsill1,Graph1}. Usually, the actual PT states are unknown, and all we can obtain are the estimated PT states and corresponding covariance information. As one of the most commonly used distance metrics, Mahalanobis distance provides an efficient way to incorporate the confidence about the PT state estimate, which is given by
\begin{align*}
	d_{k}^{i,i^{\prime}}:=(\mathbf{x}_{k}^{(i)}-\mathbf{x}_{k}^{(i^{\prime})})^{\mathrm{T}}(\mathbf{P}_{k}^{(i)}+\mathbf{P}_{k}^{(i^{\prime})})^{-1}(\mathbf{x}_{k}^{(i)}-\mathbf{x}_{k}^{(i^{\prime})}),
\end{align*}
where $\mathbf{P}_{k}^{(i)}$ is the  covariance of $\mathbf{x}_{k}^{(i)}$. Let 
\begin{align*}
	\breve{\mathbf{P}}_{k-1,\tilde{\Lambda}_{\underline{\mathbf{g}}_{k}}(j)}&:=\frac{1}{|\tilde{\Lambda}_{\underline{\mathbf{g}}_{k}}(j)|}\sum_{i\in\tilde{\Lambda}_{\underline{\mathbf{g}}_{k}}(j)}\mathbf{P}_{k-1}^{(i)},
\end{align*}
denote the average covariance of these survival PTs in group $j$. Then, we use the Mahalanobis distance between a PT state $\mathbf{x}_{k}^{(i)}$ and the virtual leader $\breve{\mathbf{x}}_{k-1,\tilde{\Lambda}_{\underline{\mathbf{g}}_{k}}(j)}$ to define a quantity $P_{i,j}\in\left[0,1\right]$ as follows:
\begin{align}\label{Pij}
	P_{i,j}:=
	\begin{cases}
		P_{0}, &\begin{array}{l}\text{if}\  r_{k-1}^{(i)}=0
		\end{array} \\
		\exp(-\frac{d_{k-1}^{i,j}}{2}), &\begin{array}{l}\text{otherwise},
		\end{array} 
	\end{cases}
\end{align}
where $P_{0}\in\left[0, 1\right]$ is a small number meaning that dividing nonexistent PTs into a group has a small probability. We define a scoring function for evaluating the group partition $\underline{\mathbf{g}}_{k}$ with given $\mathbf{x}_{k-1}$ and $\mathbf{r}_{k-1}$ as 
\begin{align*}
	\begin{split}
		s(\underline{\mathbf{g}}_{k}|\mathbf{x}_{k-1},\mathbf{r}_{k-1}):=\prod_{i\in\Lambda_{\underline{\mathbf{g}}_{k}}}P_{i,\underline{g}_{k}^{(i)}}\prod_{j\in \{1,\ldots,N(\underline{\mathbf{g}}_{k})\}\backslash\underline{g}_{k}^{(i)}}(1-P_{i,j}),
	\end{split}
\end{align*}
where $\Lambda_{\underline{\mathbf{g}}_{k}}:=\{1,\ldots,n_k\}\backslash\Lambda_{\underline{\mathbf{g}}_{k}}(0)$ is the index set of all confirmed legacy PTs at time $k$. Thus, we can define a pseudo group structure transition pmf as
\begin{align}\label{GS-pdf}
	p(\underline{\mathbf{g}}_{k}|\mathbf{x}_{k-1},\mathbf{r}_{k-1}):= \frac{s(\underline{\mathbf{g}}_{k}|\mathbf{x}_{k-1},\mathbf{r}_{k-1})}{\sum_{\underline{\mathbf{g}}_{k}^{\prime}\in\underline{\mathcal{G}}_k}s(\underline{\mathbf{g}}_{k}^{\prime}|\mathbf{x}_{k-1},\mathbf{r}_{k-1})}.
\end{align}

\subsection{Conditional pdf $p(\mathbf{z}_{k}, \mathbf{a}_{k},m_{k}, \overline{\mathbf{y}}_{k},\overline{\mathbf{g}}_{k} |\underline{\mathbf{y}}_{k},\underline{\mathbf{g}}_{k})$}
Next, we introduce the calculation of the conditional pdf $p(\mathbf{z}_{k}, \mathbf{a}_{k},m_{k}, \overline{\mathbf{y}}_{k},\overline{\mathbf{g}}_{k} |\underline{\mathbf{y}}_{k},\underline{\mathbf{g}}_{k})$ in (\ref{condition-pdf}). It is commonly assumed that given $\mathbf{y}_k$ and $\mathbf{a}_k$, the measurement vector $\mathbf{z}_{k}$ is independent of $\mathbf{g}_k$. According to the definition of $\overline{\mathbf{g}}_{k}$, it is a deterministic zero vector to represent the group structure of the new PTs. We assume that given $\underline{\mathbf{y}}_k$, the association vector $\mathbf{a}_{k}$ and the augmented new PT states $\overline{\mathbf{y}}_{k}$ are independent of $\mathbf{g}_k$. By the chain rule and the conditional independence assumption, we have
\begin{align}\label{p24}
	\begin{split}
		p(\mathbf{z}_{k}, \mathbf{a}_{k},m_{k}, \overline{\mathbf{y}}_{k},\overline{\mathbf{g}}_{k} |\underline{\mathbf{y}}_{k},\underline{\mathbf{g}}_{k})&=p(\mathbf{z}_{k}, \mathbf{a}_{k},m_{k}, \overline{\mathbf{x}}_{k},\overline{\mathbf{r}}_{k},\overline{\mathbf{g}}_{k} |\underline{\mathbf{x}}_{k},\underline{\mathbf{r}}_{k},\underline{\mathbf{g}}_{k})
		\\
		&=p(\mathbf{z}_{k}| \mathbf{a}_{k},m_{k}, \overline{\mathbf{x}}_{k},\overline{\mathbf{r}}_{k},\overline{\mathbf{g}}_{k},\underline{\mathbf{x}}_{k},\underline{\mathbf{r}}_{k},\underline{\mathbf{g}}_{k})p( \mathbf{a}_{k},m_{k}, \overline{\mathbf{x}}_{k},\overline{\mathbf{r}}_{k},\overline{\mathbf{g}}_{k}|\underline{\mathbf{x}}_{k},\underline{\mathbf{r}}_{k},\underline{\mathbf{g}}_{k})
		\\
		&=p(\mathbf{z}_{k}| \mathbf{a}_{k},m_{k}, \overline{\mathbf{x}}_{k},\overline{\mathbf{r}}_{k},\underline{\mathbf{x}}_{k},\underline{\mathbf{r}}_{k})p( \mathbf{a}_{k},m_{k}, \overline{\mathbf{x}}_{k},\overline{\mathbf{r}}_{k}|\underline{\mathbf{x}}_{k},\underline{\mathbf{r}}_{k}).
	\end{split}
\end{align}
Let $f(\mathbf{z}_{k}^{(m)}|\underline{\mathbf{x}}_{k}^{(i)})$  denote the pdf of the measurement $\mathbf{z}_{k}^{(m)}$ conditioned on the legacy PT state $\underline{\mathbf{x}}_{k}^{(i)}$. The target-originated measurements are assumed conditionally independent of each other and also conditionally independent of all clutters. Moreover, the number of clutters is assumed Poisson distributed with a mean $\mu_{\mathrm{c}}$, and the clutters are assumed independent and identically distributed with pdf $f_{\mathrm{c}}(\mathbf{z}_{k}^{(m)})$ \cite{BP-MTT1,BP-MTT2}. Then, the pdf $p(\mathbf{z}_{k}| \mathbf{a}_{k},m_{k}, \overline{\mathbf{x}}_{k},\overline{\mathbf{r}}_{k},\underline{\mathbf{x}}_{k},\underline{\mathbf{r}}_{k})$ is given by
\begin{align}\label{pz0}
	\begin{split}
		p(\mathbf{z}_{k}| \mathbf{a}_{k},m_{k}, \overline{\mathbf{x}}_{k},\overline{\mathbf{r}}_{k},\underline{\mathbf{x}}_{k},\underline{\mathbf{r}}_{k})=\big(\prod_{m=1}^{m_{k}} f_{\mathrm{c}}(\mathbf{z}_{k}^{(m)})\big)\big(\prod_{i \in \mathcal{D}_{\mathbf{a}_{k}}}\frac{f(\mathbf{z}_{k}^{(a_{k}^{(i)})} | \underline{\mathbf{x}}_{k}^{(i)})}{f_{\mathrm{c}}(\mathbf{z}_{k}^{(a_{k}^{(i)})})}\big)\big(\prod_{m^{\prime} \in \mathcal{I}_{\overline{\mathbf{r}}_{k}}}\frac{f(\mathbf{z}_{k}^{(m^{\prime})} | \overline{\mathbf{x}}_{k}^{(m^{\prime})})}{f_{\mathrm{c}}(\mathbf{z}_{k}^{(m^{\prime})})}\big),
	\end{split}
\end{align} 
where $\mathcal{D}_{\mathbf{a}_{k}}$ and $\mathcal{I}_{\overline{\mathbf{r}}_{k}}$ are the index sets of detected legacy PTs and new PTs at time $k$, respectively,
\begin{align*}
	&\mathcal{D}_{\mathbf{a}_{k}}:=\{i\in \{1,\ldots,n_{k}\}: \underline{r}_k^{(i)}=1, a_{k}^{(i)}\neq 0\},
	\\
	&\mathcal{I}_{\overline{\mathbf{r}}_{k}}:=\{m\in \{1,\ldots,m_{k}\}: \overline{r}_k^{(m)}=1\}.
\end{align*} 
Since the new PTs and the legacy PTs at time $k$ are grouped at the next time instance, we assume that the new PTs are independent of the legacy PTs at time $k$. Then, the pdf $p( \mathbf{a}_{k},m_{k}, \overline{\mathbf{x}}_{k},\overline{\mathbf{r}}_{k}|\underline{\mathbf{x}}_{k},\underline{\mathbf{r}}_{k})$ is obtained as
\begin{align}\label{pz1}
	\begin{split}
		p( \mathbf{a}_{k},m_{k}, \overline{\mathbf{x}}_{k},\overline{\mathbf{r}}_{k}|\underline{\mathbf{x}}_{k},\underline{\mathbf{r}}_{k})&=p(\overline{\mathbf{x}}_{k}|\mathbf{a}_{k},\overline{\mathbf{r}}_{k},m_{k},\underline{\mathbf{x}}_{k},\underline{\mathbf{r}}_{k})p( \mathbf{a}_{k},\overline{\mathbf{r}}_{k},m_{k}|\underline{\mathbf{x}}_{k},\underline{\mathbf{r}}_{k})
		\\
		&=p(\overline{\mathbf{x}}_{k}|\overline{\mathbf{r}}_{k},m_{k})p( \mathbf{a}_{k},\overline{\mathbf{r}}_{k},m_{k}|\underline{\mathbf{x}}_{k},\underline{\mathbf{r}}_{k}),
	\end{split}
\end{align}
with
\begin{align*}
	p(\overline{\mathbf{x}}_{k}|\overline{\mathbf{r}}_{k},m_{k})=\big(\prod_{m\in\mathcal{I}_{\overline{\mathbf{r}}_{k}}} f_{\mathrm{b}}(\overline{\mathbf{x}}_{k}^{(m)})\big)\prod_{m^{\prime}\notin\mathcal{I}_{\overline{\mathbf{r}}_{k}}\atop m^{\prime}\in \{1,\ldots,m_{k}\}} f_{\mathrm{d}}(\overline{\mathbf{x}}_{k}^{(m^{\prime})}),
\end{align*}
where $f_{\mathrm{b}}(\overline{\mathbf{x}}_{k}^{(m)})$ is a prior pdf for the new PTs. The number of the new PTs at time $k$
is assumed Poisson distributed with a
mean $\mu_{\mathrm{b}}$, which is independent of the number of legacy PTs and of the number of clutters. Let $p_{\mathrm{d}}(\underline{\mathbf{x}}_{k}^{(i)})$ denote the probability of the legacy PT $\underline{\mathbf{x}}_{k}^{(i)}$ detected by sensor at time $k$, that is, with the probability $1-p_{\mathrm{d}}(\underline{\mathbf{x}}_{k}^{(i)})$ of misdetection. The prior pmf of $\mathbf{a}_{k}$, $\overline{\mathbf{r}}_{k}$ and $m_{k}$ conditioned on $\underline{\mathbf{x}}_{k}$, $\underline{\mathbf{r}}_{k}$ is given by
\begin{align}\label{pz3}
	\begin{split}
		p( \mathbf{a}_{k},\overline{\mathbf{r}}_{k},m_{k}|\underline{\mathbf{x}}_{k},\underline{\mathbf{r}}_{k})&=\frac{1}{m_{k} !}e^{-\mu_{\mathrm{b}}}(\mu_{\mathrm{b}})^{|\mathcal{I}_{\overline{\mathbf{r}}_{k}}|}e^{-\mu_{\mathrm{c}}}(\mu_{\mathrm{c}})^{m_{k}-|\mathcal{D}_{\mathbf{a}_{k}}|-|\mathcal{I}_{\overline{\mathbf{r}}_{k}}|}\psi(\mathbf{a}_{k})\big(\prod_{m\in\mathcal{I}_{\overline{\mathbf{r}}_{k}}}\Gamma_{\mathbf{a}_{k}}^{(m)}\big)
		\\
		&\quad\times \big(\prod_{i\in\mathcal{D}_{\mathbf{a}_{k}}}\underline{r}_k^{(i)}p_{\mathrm{d}}(\underline{\mathbf{x}}_{k}^{(i)})\big)\prod_{i^{\prime} \notin \mathcal{D}_{\mathbf{a}_{k}}\atop i^{\prime}\in \{1,\ldots,n_{k}\}}(1-\underline{r}_k^{(i^{\prime})}p_{\mathrm{d}}(\underline{\mathbf{x}}_{k}^{(i^{\prime})})),
	\end{split}
\end{align}
with
\begin{align*}
	\psi(\mathbf{a}_{k})&:=
	\begin{cases}
		0, &\begin{array}{l}\exists\ i,\ i^{\prime}\in \{1,\ldots,n_{k}\}, \text { such that } i\neq i^{\prime}  \text { and } a_{k}^{(i)}=a_{k}^{(i^{\prime})}\neq0
		\end{array} \\
		1, &\begin{array}{l}\text{otherwise, }\end{array}
	\end{cases}
	\\
	\Gamma_{\mathbf{a}_{k}}^{(m)}&:=
	\begin{cases}
		0, &\begin{array}{l}\exists\ i \in \{1,\ldots,n_{k}\}, \text { such that } a_{k}^{(i)}=m
		\end{array} \\
		1, &\begin{array}{l}\text{otherwise, }\end{array}
	\end{cases}
\end{align*}
where the indicator functions $\psi(\mathbf{a}_{k})$ and $\Gamma_{\mathbf{a}_{k}}^{(m)}$ ensure that each measurement can only be associated once, either with a PT or with clutter. According to (\ref{pz0})-(\ref{pz3}), the pdf in (\ref{p24}) can be written as
\begin{align}\label{pl}
	\begin{split}
		p(\mathbf{z}_{k}, \mathbf{a}_{k},m_{k}, \overline{\mathbf{y}}_{k},\overline{\mathbf{g}}_{k} |\underline{\mathbf{y}}_{k},\underline{\mathbf{g}}_{k})&\propto\psi(\mathbf{a}_{k})\big(\prod_{i=1}^{n_{k}}q(\underline{\mathbf{x}}_{k}^{(i)}, \underline{r}_{k}^{(i)}, a_{k}^{(i)}; \mathbf{z}_k)\big)\prod_{m=1}^{m_{k}}v_1(\overline{\mathbf{x}}_{k}^{(m)}, \overline{r}_{k}^{(m)}, \mathbf{a}_{k})v_2(\overline{\mathbf{x}}_{k}^{(m)}, \overline{r}_{k}^{(m)}; \mathbf{z}_{k}),
	\end{split}
\end{align}
where $q(\underline{\mathbf{x}}_{k}^{(i)}, \underline{r}_{k}^{(i)}, a_{k}^{(i)}; \mathbf{z}_k)$ is defined as
\begin{align*}
	\begin{split}
		q(\underline{\mathbf{x}}_{k}^{(i)}, 1, a_{k}^{(i)}; \mathbf{z}_k):=
		\begin{cases}
			\frac{p_{\mathrm{d}}(\underline{\mathbf{x}}_{k}^{(i)})f(\mathbf{z}_{k}^{(a_{k}^{(i)})}| \underline{\mathbf{x}}_{k}^{(i)})}{\mu_{\mathrm{c}}f_{\mathrm{c}}(\mathbf{z}_{k}^{(a_{k}^{(i)})})}, &\begin{array}{l}a_{k}^{(i)}\neq0
			\end{array} \\
			1-p_{\mathrm{d}}(\underline{\mathbf{x}}_{k}^{(i)}), &\begin{array}{l}a_{k}^{(i)}=0,\end{array}
		\end{cases}
	\end{split}
\end{align*} 
with $q(\underline{\mathbf{x}}_{k}^{(i)}, 0, a_{k}^{(i)}; \mathbf{z}_k):=\mathrm{I}(a_{k}^{(i)})$, and $v_1(\overline{\mathbf{x}}_{k}^{(m)}, \overline{r}_{k}^{(m)}, \mathbf{a}_{k})$ is defined as
\begin{align}\label{v1}
	\begin{split}
		v_1(\overline{\mathbf{x}}_{k}^{(m)}, 1, \mathbf{a}_{k}):=
		\begin{cases}
			0, &\begin{array}{l}\exists\ i \in \{1,\ldots,n_{k}\},\ \text {such that } a_{k}^{(i)}=m
			\end{array} \\
			\frac{\mu_{\mathrm{b}}}{\mu_{\mathrm{c}}}f_{\mathrm{b}}(\overline{\mathbf{x}}_{k}^{(m)}), &\begin{array}{l}\text{otherwise,}\end{array}
		\end{cases}
	\end{split}
\end{align}
with 
$v_1(\overline{\mathbf{x}}_{k}^{(m)}, 0, \mathbf{a}_{k}):=f_{\mathrm{d}}(\overline{\mathbf{x}}_{k}^{(m)})$, and 
\begin{align}\label{v2}
	\begin{split}
		v_2(\overline{\mathbf{x}}_{k}^{(m)}, \overline{r}_{k}^{(m)}; \mathbf{z}_{k}):=
		\begin{cases}
			\frac{f(\mathbf{z}_{k}^{(m)}| \overline{\mathbf{x}}_{k}^{(m)})}{f_{\mathrm{c}}(\mathbf{z}_{k}^{(m)})}, &\begin{array}{l}\overline{r}_{k}^{(m)}=1
			\end{array} \\
			1, &\begin{array}{l}\overline{r}_{k}^{(m)}=0.\end{array}
		\end{cases}
	\end{split}
\end{align}
Consequently, substituting (\ref{ps}) and (\ref{pl}) into (\ref{pdf}), the joint posterior pdf $p(\mathbf{y}_{1:k}, \mathbf{g}_{1:k}, \mathbf{a}_{1:k}|\mathbf{z}_{1:k})$ is factorized as 
\begin{align}\label{p1}
	\begin{split}
		p(\mathbf{y}_{1:k}, \mathbf{g}_{1:k}, \mathbf{a}_{1:k}|\mathbf{z}_{1:k})&\propto\prod_{k^{\prime}=1}^{k}p(\underline{\mathbf{y}}_{k^{\prime}},\underline{\mathbf{g}}_{k^{\prime}}|\mathbf{y}_{k^{\prime}-1},\mathbf{g}_{k^{\prime}-1})\big(\prod_{i=1}^{n_{k^{\prime}}}q(\underline{\mathbf{x}}_{k^{\prime}}^{(i)}, \underline{r}_{k^{\prime}}^{(i)}, a_{k^{\prime}}^{(i)}; \mathbf{z}_{k}^{\prime})\big)
		\\
		&\quad\times\psi(\mathbf{a}_{k^{\prime}})\prod_{m=1}^{m_{k^{\prime}}}v_1(\overline{\mathbf{x}}_{k^{\prime}}^{(m)}, \overline{r}_{k^{\prime}}^{(m)}, \mathbf{a}_{k^{\prime}})v_2(\overline{\mathbf{x}}_{k^{\prime}}^{(m)}, \overline{r}_{k^{\prime}}^{(m)}; \mathbf{z}_{k^{\prime}}).
	\end{split}
\end{align}
\begin{remark}
	As shown in the factorization (\ref{p1}) of the joint posterior pdf, the group structure not only helps to model the evolution of targets, but also has an important impact on the data association (i.e., the likelihood calculation). Therefore, it is crucial to consider the group structure in the GTT problem.
\end{remark}
\section{The Proposed GTBP Method}
In this section, we derive a further factorization of the joint posterior pdf $p(\mathbf{y}_{1:k},\mathbf{g}_{1:k}, \mathbf{a}_{1:k}|\mathbf{z}_{1:k})$ by stretching the factor $\psi(\mathbf{a}_{k})$, and then propose the GTBP method.
\subsection{Factor Stretching and Joint Posterior pdf}
\begin{figure}[hbtp]
	\centering 
	\includegraphics[width=0.8\linewidth]{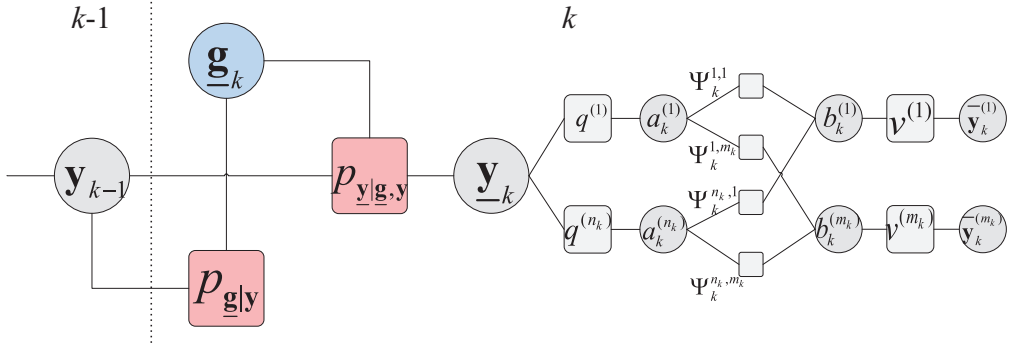}
	\caption{The factor graph description of the factorization (\ref{p2}) for GTT, shown for the time $k$. Some abbreviations are used: $p_{\underline{\mathbf{g}}|\mathbf{y}}:=p(\underline{\mathbf{g}}_{k}|\mathbf{x}_{k-1}, \mathbf{r}_{k-1})$, $p_{\underline{\mathbf{y}}|\underline{\mathbf{g}},\mathbf{y}}:=p(\underline{\mathbf{x}}_{k}, \underline{\mathbf{r}}_{k}|\underline{\mathbf{g}}_{k}, \mathbf{x}_{k-1}, \mathbf{r}_{k-1})$,  $q^{(i)}:=q(\underline{\mathbf{x}}_{k}^{(i)}, \underline{r}_{k}^{(i)}, a_{k}^{(i)}; \mathbf{z}_{k})$, $v^{(m)}:=v(\overline{\mathbf{x}}_{k}^{(m)}, \overline{r}_{k}^{(m)}, b_{k}^{(m)}; \mathbf{z}_{k}^{(m)})$. 
	}
	\label{figure2}
\end{figure}

Note that in (\ref{p1}), we factorize the joint posterior pdf $p(\mathbf{y}_{1:k}, \mathbf{g}_{1:k}, \mathbf{a}_{1:k}|\mathbf{z}_{1:k})$ into some products. However, the factor  $\psi(\mathbf{a}_{k})$ is a coupled function of all entries of the target-oriented association vector $\mathbf{a}_{k}$, which may suffer high-dimensional discrete marginalizations when using BP to compute the messages. To avoid this, the stretching principle in factor graphs can be applied. Following \cite{FG-BP,BP-MTT1,BP-MTT2}, we consider introducing the measurement-oriented association vector $\mathbf{b}_{k}:=\left[b_{k}^{(1)},\ldots,b_{k}^{(m_{k})}\right]^{\mathrm{T}}$ with 
\begin{align*}
	b_{k}^{(m)}:=
	\begin{cases}
		i \in\left\{1, \ldots, n_{k}\right\}, &\begin{array}{l}\text { if the  measurement } \mathbf{z}_k^{(m)}  \text { is } \text{generated by } \underline{\mathbf{x}}_k^{(i)}
		\end{array} \\
		0, &\begin{array}{l}\text { if } \mathbf{z}_k^{(m)}  \text { is not generated } \text { by a legacy PT}.
		\end{array} \\
	\end{cases}
\end{align*}
Notably, the measurement-oriented association vector $\mathbf{b}_{k}$ is redundant with $\mathbf{a}_{k}$, that is, one of the two association vectors is determined and the other is determined as well. By introducing $\mathbf{b}_{k}$, the factor $\psi(\mathbf{a}_{k})$ can be stretched and equivalently replaced by 
\begin{align*}
	\psi(\mathbf{a}_{k},\mathbf{b}_{k}):=\prod_{i=1}^{n_{k}}\prod_{m=1}^{m_{k}}\Psi_{k}^{i,m}(a_{k}^{(i)},b_{k}^{(m)}),
\end{align*}
where 
\begin{align*}
	\Psi_{k}^{i,m}(a_{k}^{(i)},b_{k}^{(m)}):=
	\begin{cases}
		0, &\begin{array}{l}a_{k}^{(i)}=m,\  b_{k}^{(m)}\neq i \text{ or } b_{k}^{(m)}=i,\  a_{k}^{(i)}\neq m
		\end{array} \\
		1, &\begin{array}{l}\text{otherwise. }\end{array}
	\end{cases}
\end{align*}
Hereafter, we abbreviate $\Psi_{k}^{i,m}(a_{k}^{(i)},b_{k}^{(m)})$ as $\Psi_{k}^{i,m}$ for notational convenience.
Note that in (\ref{v1}), the condition that there exists $i \in \{1,\ldots,n_{k}\}$ such that $a_{k}^{(i)}=m$ is equal to the condition that $b_{k}^{(m)}\in\{1,\ldots,n_{k}\}$. According to the definitions (\ref{v1})-(\ref{v2}) of $v_1(\overline{\mathbf{x}}_{k}^{(m)}, \overline{r}_{k}^{(m)}, \mathbf{a}_{k})$ and $v_2(\overline{\mathbf{x}}_{k}^{(m)}, \overline{r}_{k}^{(m)}; \mathbf{z}_{k})$, one can easily verify that their product can be replaced by $v(\overline{\mathbf{x}}_{k}^{(m)}, \overline{r}_{k}^{(m)}, b_{k}^{(m)}; \mathbf{z}_{k}^{(m)})$,
\begin{align}\label{def-v}
	\begin{split}
		v(\overline{\mathbf{x}}_{k}^{(m)}, 1, b_{k}^{(m)}; \mathbf{z}_{k}^{(m)}):=\begin{cases}
			0, &\begin{array}{l}b_{k}^{(m)}\in\{1,\ldots,n_{k}\}
			\end{array} \\
			\frac{\mu_{\mathrm{b}}f_{\mathrm{b}}(\overline{\mathbf{x}}_{k}^{(m)})f(\mathbf{z}_{k}^{(m)}| \overline{\mathbf{x}}_{k}^{(m)})}{\mu_{\mathrm{c}}f_{\mathrm{c}}(\mathbf{z}_{k}^{(m)})}, &\begin{array}{l}b_{k}^{(m)}=0,\end{array}
		\end{cases}
	\end{split}
\end{align}
with 
$v(\overline{\mathbf{x}}_{k}^{(m)}, 0, b_{k}^{(m)}; \mathbf{z}_{k}^{(m)}):=f_{\mathrm{d}}(\overline{\mathbf{x}}_{k}^{(m)})$. Let $\mathbf{b}_{1:k}$ denote the stacked measurement-oriented association vector from time 1 to time $k$. Thus, we can further factorize the joint posterior pdf $p(\mathbf{y}_{1:k}, \mathbf{g}_{1:k}, \mathbf{a}_{1:k}, \mathbf{b}_{1:k}|\mathbf{z}_{1:k})$ as follows:
\begin{align}\label{p2}
	\begin{split}
		p(\mathbf{y}_{1:k},\mathbf{g}_{1:k}, \mathbf{a}_{1:k}, \mathbf{b}_{1:k}|\mathbf{z}_{1:k})&\propto\prod_{k^{\prime}=1}^{k}\bigg(p(\underline{\mathbf{y}}_{k^{\prime}},\underline{\mathbf{g}}_{k^{\prime}}|\mathbf{y}_{k^{\prime}-1},\mathbf{g}_{k^{\prime}-1})\big(\prod_{i=1}^{n_{k^{\prime}}}q(\underline{\mathbf{x}}_{k^{\prime}}^{(i)}, \underline{r}_{k^{\prime}}^{(i)}, a_{k^{\prime}}^{(i)}; \mathbf{z}_{k^{\prime}})\prod_{m=1}^{m_{k^{\prime}}}\Psi_{k^{\prime}}^{i,m}\big)
		\\ 
		&\quad \times\prod_{m^{\prime}=1}^{m_{k^{\prime}}}v(\overline{\mathbf{x}}_{k^{\prime}}^{(m^{\prime})}, \overline{r}_{k^{\prime}}^{(m^{\prime})}, b_{k^{\prime}}^{(m^{\prime})}; \mathbf{z}_{k^{\prime}}^{(m^{\prime})})\bigg).
	\end{split}
\end{align}
A factor graph representation for this factorization is shown in Fig. \ref{figure2}, which is mainly depicted for the time $k$. Herein, the factor nodes and variable nodes are drawn with squares and circles, respectively.
\subsection{The GTBP Method}
Based on the factorization (\ref{p2}) and the devised factor graph in Fig. \ref{figure2}, we calculate the beliefs in detail via the message passing scheme (\ref{ftv}), and then obtain the desired marginal posterior pdfs and pmfs.
\subsubsection{Prediction of Group Structure and Target State}
First, the message $\alpha_k(\underline{\mathbf{g}}_{k})$ passed from the factor node $p(\underline{\mathbf{g}}_{k}|\mathbf{x}_{k-1},\mathbf{r}_{k-1})$ to the variable node $\underline{\mathbf{g}}_{k}$  is calculated as 
\begin{align}\label{pre-gs}
	\begin{split}
		\alpha_k(\underline{\mathbf{g}}_{k})=\sum_{\mathbf{r}_{k-1}\in\mathcal{R}_{k-1}}\int p(\underline{\mathbf{g}}_{k}|\mathbf{x}_{k-1}, \mathbf{r}_{k-1})\widetilde{p}(\mathbf{x}_{k-1},\mathbf{r}_{k-1})\mathrm{d}{\mathbf{x}_{k-1}},
	\end{split}
\end{align}
where $\widetilde{p}(\mathbf{x}_{k-1},\mathbf{r}_{k-1})$ is the approximation of the marginal posterior pdf $p(\mathbf{x}_{k-1},\mathbf{r}_{k-1}|\mathbf{z}_{1:k-1})$ obtained at time $k-1$. Then, the message $\alpha_k(\underline{\mathbf{y}}_{k})=\alpha_k(\underline{\mathbf{x}}_{k}, \underline{\mathbf{r}}_{k})$ passed from the factor node $p(\underline{\mathbf{y}}_{k}|\underline{\mathbf{g}}_{k},\mathbf{y}_{k-1})$ to the variable node $\underline{\mathbf{y}}_{k}$  is calculated as
\begin{align}\label{alpha-gs}
	\begin{split}
		\alpha_k(\underline{\mathbf{x}}_{k}, \underline{\mathbf{r}}_{k})&=\sum_{\underline{\mathbf{g}}_{k}\in\underline{\mathcal{G}}_{k}}\sum_{\mathbf{r}_{k-1}\in\mathcal{R}_{k-1}}\int\alpha_k(\underline{\mathbf{g}}_{k})\widetilde{p}(\mathbf{x}_{k-1},\mathbf{r}_{k-1}) p(\underline{\mathbf{x}}_{k}, \underline{\mathbf{r}}_{k}|\underline{\mathbf{g}}_{k}, \mathbf{x}_{k-1}, \mathbf{r}_{k-1})
		\mathrm{d}{\mathbf{x}_{k-1}}
		\\
		&=\sum_{\underline{\mathbf{g}}_{k}\in\underline{\mathcal{G}}_{k}}\alpha_k(\underline{\mathbf{g}}_{k})\alpha_k(\underline{\mathbf{x}}_{k}, \underline{\mathbf{r}}_{k}|\underline{\mathbf{g}}_{k}),
	\end{split}
\end{align}
where
\begin{align}\label{alpha1}
	\begin{split}
		\alpha_k(\underline{\mathbf{x}}_{k}, \underline{\mathbf{r}}_{k}|\underline{\mathbf{g}}_{k})=\sum_{\mathbf{r}_{k-1}\in\mathcal{R}_{k-1}}\int \widetilde{p}(\mathbf{x}_{k-1},\mathbf{r}_{k-1})p(\underline{\mathbf{x}}_{k}, \underline{\mathbf{r}}_{k}|\underline{\mathbf{g}}_{k}, \mathbf{x}_{k-1}, \mathbf{r}_{k-1})\mathrm{d}{\mathbf{x}_{k-1}}.
	\end{split}
\end{align}
\begin{remark}
	Note that in (\ref{alpha-gs}), the message $\alpha_k(\underline{\mathbf{x}}_{k}, \underline{\mathbf{r}}_{k})$ involves the weighted summation of $\alpha_k(\underline{\mathbf{x}}_{k}, \underline{\mathbf{r}}_{k}|\underline{\mathbf{g}}_{k})$ over possible group structures. That is, the evolution of targets is modeled as the co-action of the group or single-target motions under different group structures. Herein, the group structure probabilities $\alpha_k(\underline{\mathbf{g}}_{k})$, namely the target state transition mode weights, will be updated by the data association presented subsequently. Particularly, if the group structure is deterministic and each target is divided into a group, then GTBP degrades to the BP method for MTT \cite{BP-MTT2}.
\end{remark}
\subsubsection{Measurement Evaluation and Iterative Data Association}
The messages $\beta_k(a_{k}^{(i)})$ passed from the factor nodes $q(\underline{\mathbf{x}}_{k}^{(i)}, \underline{r}_{k}^{(i)}, a_{k}^{(i)}; \mathbf{z}_{k})$ to the variable nodes $a_{k}^{(i)}$ are computed by
\begin{align}\label{beta}
	\begin{split}
		\beta_k(a_{k}^{(i)})&=\sum_{\underline{\mathbf{r}}_{k}\in\underline{\mathcal{R}}_{k}}\int q(\underline{\mathbf{x}}_{k}^{(i)}, \underline{r}_{k}^{(i)}, a_{k}^{(i)}; \mathbf{z}_{k}) \alpha_k(\underline{\mathbf{x}}_{k}, \underline{\mathbf{r}}_{k})\mathrm{d}{\underline{\mathbf{x}}_{k}}.
	\end{split}
\end{align}
For the new PTs, the messages $\xi_k(b_{k}^{(m)})$ passed from the factor nodes $v(\overline{\mathbf{x}}_{k}^{(m)}, \overline{r}_{k}^{(m)}, b_{k}^{(m)}; \mathbf{z}_{k}^{(m)})$ to the variable nodes $b_{k}^{(m)}$ are computed by
\begin{align}\label{xi}
	\begin{split}
		\xi_k(b_{k}^{(m)})&=\sum_{\overline{r}_{k}^{(m)}\in\{0,1\} }\int v(\overline{\mathbf{x}}_{k}^{(m)}, \overline{r}_{k}^{(m)}, b_{k}^{(m)}; \mathbf{z}_{k}^{(m)})\mathrm{d}{\overline{\mathbf{x}}_{k}^{(m)}}
		\\
		&=\int v(\overline{\mathbf{x}}_{k}^{(m)}, 1, b_{k}^{(m)}; \mathbf{z}_{k}^{(m)})\mathrm{d}{\overline{\mathbf{x}}_{k}^{(m)}}+1.
	\end{split}
\end{align}
Once the incoming messages $\beta_k(a_{k}^{(i)})$ to the data association part have been calculated, the iterative message calculation between all the nodes $a_{k}^{(i)}$, $b_{k}^{(m)}$ and $\Psi_{k}^{i,m}$ are performed. In the iteration, the messages $\varphi_{\Psi_{k}^{i,m} \rightarrow b_{k}^{(m)}}^{\left[\ell\right]}(b_{k}^{(m)})$ and $\varphi_{\Psi_{k}^{i,m} \rightarrow a_{k}^{(i)}}^{\left[\ell\right]}(a_{k}^{(i)})$ are updated by
\begin{align}
	\begin{split}\label{message-beta}
		&\varphi_{\Psi_{k}^{i,m} \rightarrow b_{k}^{(m)}}^{\left[\ell\right]}(b_{k}^{(m)})=\sum_{a_{k}^{(i)}=0}^{m_{k}}\beta_k(a_{k}^{(i)})\Psi_{k}^{i,m}\prod_{{m^{\prime}=1}\atop{m^{\prime}\neq m}}^{m_{k}}\varphi_{\Psi_{k}^{i,m^{\prime}} \rightarrow a_{k}^{(i)}}^{\left[\ell\right]}(a_{k}^{(i)}),
	\end{split}
	\\
	\begin{split}\label{message-xi}
		&\varphi_{\Psi_{k}^{i,m} \rightarrow a_{k}^{(i)}}^{\left[\ell\right]}(a_{k}^{(i)})=\sum_{b_{k}^{(m)}=0}^{n_{k}}\xi_k(b_{k}^{(m)}) \Psi_{k}^{i,m}\prod_{{i^{\prime}=1}\atop{i^{\prime}\neq i}}^{n_{k}}\varphi_{\Psi_{k}^{i^{\prime},m} \rightarrow b_{k}^{(m)}}^{\left[\ell-1\right]}(b_{k}^{(m)}),
	\end{split}
\end{align}
where the subscript $\ell$ denotes the number of iteration, $i=1,\ldots,n_{k}$ and $m=1,\ldots,m_{k}$. The iterative loop of (\ref{message-beta})-(\ref{message-xi}) is initialized by setting
\begin{align}\label{ini-beta}
	\varphi_{\Psi_{k}^{i,m} \rightarrow b_{k}^{(m)}}^{\left[0\right]}(b_{k}^{(m)})=&\sum_{a_{k}^{(i)}=0}^{m_{k}}\beta_k(a_{k}^{(i)}) \Psi_{k}^{i,m}.
\end{align}
An efficient implementation of Matlab code for this iteration is provided in \cite{Williams3}. The iteration of (\ref{message-beta})-(\ref{message-xi}) terminates when meeting the max number of iteration or the Frobenius norm of the beliefs between two consecutive iterations is less than a certain threshold. We denote the number of iterations when meeting the stopping criteria as $\ell_k$, then the messages passed from $a_{k}^{(i)}$ to $q(\underline{\mathbf{x}}_{k}^{(i)}, \underline{r}_{k}^{(i)}, a_{k}^{(i)}; \mathbf{z}_{k})$ are obtained as
\begin{align}\label{kappa}
	\kappa_k(a_{k}^{(i)})=	\prod_{m=1}^{m_{k}}\varphi_{\Psi_{k}^{i,m} \rightarrow a_{k}^{(i)}}^{\left[\ell_k\right]}(a_{k}^{(i)}),
\end{align}
for $i=1,\ldots,n_{k}$, and the  the messages passed from $b_{k}^{(m)}$ to $v(\overline{\mathbf{x}}_{k}^{(m)}, \overline{r}_{k}^{(m)}, b_{k}^{(m)}; \mathbf{z}_{k}^{(m)})$ are obtained as
\begin{align}\label{iota}
	\iota_k(b_{k}^{(m)})=	\prod_{i=1}^{n_{k}}\varphi_{\Psi_{k}^{i,m} \rightarrow b_{k}^{(m)}}^{\left[\ell_k\right]}(b_{k}^{(m)}).
\end{align}
\subsubsection{Measurement Update and Belief Calculation}\label{cite_sub}
When the messages $\kappa_k(a_{k}^{(i)})$ are obtained, we can calculate the messages $\gamma_k^{(i)}(\underline{\mathbf{x}}_k^{(i)}, \underline{r}_k^{(i)})$ for the legacy PTs, which are passed from $q(\underline{\mathbf{x}}_{k}^{(i)}, \underline{r}_{k}^{(i)}, a_{k}^{(i)}; \mathbf{z}_{k})$ to $\underline{\mathbf{y}}_k^{(i)}$,
\begin{align}\label{gamma_k}
	\begin{split}
		\gamma_k^{(i)}(\underline{\mathbf{x}}_k^{(i)}, 1)=\sum_{a_{k}^{(i)}=0}^{m_{k}}  q(\underline{\mathbf{x}}_{k}^{(i)}, 1, a_{k}^{(i)}; \mathbf{z}_{k}) \kappa_k(a_{k}^{(i)}),
	\end{split}
\end{align}
with $\gamma_k^{(i)}(\underline{\mathbf{x}}_k^{(i)}, 0)=\kappa_k(0)$. As a consequence, the belief $\widetilde{p}(\underline{\mathbf{y}}_{k})=\widetilde{p}(\underline{\mathbf{x}}_{k}, \underline{\mathbf{r}}_{k})$ approximating the marginal posterior pdf $p(\underline{\mathbf{x}}_{k}, \underline{\mathbf{r}}_{k}|\mathbf{z}_{1:k})$ is calculated as
\begin{align}\label{app-pdf}
	\begin{split}
		\widetilde{p}(\underline{\mathbf{x}}_{k}, \underline{\mathbf{r}}_{k})&=\frac{1}{C(\underline{\mathbf{x}}_{k})} \alpha_k(\underline{\mathbf{x}}_{k},\underline{\mathbf{r}}_{k})\prod_{i=1}^{n_{k}}\gamma_k^{(i)}(\underline{\mathbf{x}}_k^{(i)}, \underline{r}_k^{(i)})
		\\
		&=\frac{1}{C(\underline{\mathbf{x}}_{k})}\sum_{\underline{\mathbf{g}}_{k}\in\underline{\mathcal{G}}_{k}}\big(\alpha_k(\underline{\mathbf{g}}_{k})\alpha_k(\underline{\mathbf{x}}_{k}, \underline{\mathbf{r}}_{k}|\underline{\mathbf{g}}_{k})\prod_{i=1}^{n_{k}}\gamma_k^{(i)}(\underline{\mathbf{x}}_k^{(i)}, \underline{r}_k^{(i)})\big),
	\end{split}
\end{align}
where $C(\underline{\mathbf{x}}_{k})$ is a normalization constant such that $\sum_{\underline{\mathbf{r}}_{k}\in\underline{\mathcal{R}}_{k}}\int\widetilde{p}(\underline{\mathbf{x}}_{k},\underline{\mathbf{r}}_{k})\mathrm{d}{\underline{\mathbf{x}}_{k}}=1$. 

For the new PTs, the messages $\varsigma_k^{(m)}(\overline{\mathbf{x}}_k^{(m)}, \overline{r}_k^{(m)})$ passed from $v(\overline{\mathbf{x}}_{k}^{(m)}, \overline{r}_{k}^{(m)}, b_{k}^{(m)}; \mathbf{z}_{k}^{(m)})$ to $\overline{\mathbf{y}}_k^{(m)}$ are given by
\begin{align}\label{varsigma}
	\begin{split}
		\varsigma_k^{(m)}(\overline{\mathbf{x}}_k^{(m)}, 1)&=\sum_{b_{k}^{(m)}=0}^{n_{k}}  v(\overline{\mathbf{x}}_{k}^{(m)}, 1, b_{k}^{(m)}; \mathbf{z}_{k}^{(m)}) \iota_k(b_{k}^{(m)})
		\\
		&=\frac{\mu_{\mathrm{b}}f_{\mathrm{b}}(\overline{\mathbf{x}}_{k}^{(m)})f(\mathbf{z}_{k}^{(m)}| \overline{\mathbf{x}}_{k}^{(m)})}{\mu_{\mathrm{c}}f_{\mathrm{c}}(\mathbf{z}_{k}^{(m)})}\iota_k(0),
	\end{split}
\end{align}
with $\varsigma_k^{(m)}(\overline{\mathbf{x}}_k^{(m)}, 0)=\sum_{b_{k}^{(m)}=0}^{n_{k}} \iota_k(b_{k}^{(m)})f_{\mathrm{d}}(\overline{\mathbf{x}}_{k}^{(m)})$. Then, the belief $\widetilde{p}(\overline{\mathbf{y}}_{k}^{(m)})=\widetilde{p}(\overline{\mathbf{x}}_{k}^{(m)}, \overline{r}_{k}^{(m)})$ approximating the marginal posterior pdf $p(\overline{\mathbf{x}}_{k}^{(m)}, \overline{r}_{k}^{(m)}|\mathbf{z}_{1:k})$ is calculated by
\begin{align}\label{new-pt}
	\widetilde{p}(\overline{\mathbf{x}}_{k}^{(m)},\overline{r}_{k}^{(m)})&=\frac{1}{C(\overline{\mathbf{x}}_k^{(m)})} \varsigma_k^{(m)}(\overline{\mathbf{x}}_k^{(m)}, \overline{r}_{k}^{(m)}),
\end{align}
where $C(\overline{\mathbf{x}}_k^{(m)})$ is a normalization constant such that $\sum_{\overline{r}_{k}^{(m)}\in\{0,1\}}\int\widetilde{p}(\overline{\mathbf{x}}_{k}^{(m)},\overline{r}_{k}^{(m)})\mathrm{d}{\overline{\mathbf{x}}_{k}^{(m)}}=1$. 

Furthermore, the beliefs $\widetilde{p}( \underline{\mathbf{g}}_{k})$ approximating the marginal
posterior pmfs $p(\underline{\mathbf{g}}_{k}|\mathbf{z}_{1:k})$ are obtained as
\begin{align}\label{pdf-g}
	\begin{split}
		\widetilde{p}( \underline{\mathbf{g}}_{k})&=\frac{1}{C(\underline{\mathbf{x}}_{k})}\alpha_k(\underline{\mathbf{g}}_{k})\sum_{\underline{\mathbf{r}}_{k}\in\underline{\mathcal{R}}_k}\int\alpha_k(\underline{\mathbf{x}}_{k}, \underline{\mathbf{r}}_{k}|\underline{\mathbf{g}}_{k})\prod_{i=1}^{n_{k}}\gamma_k^{(i)}(\underline{\mathbf{x}}_k^{(i)}, \underline{r}_k^{(i)})\mathrm{d}{\underline{\mathbf{x}}_{k}}.
	\end{split}
\end{align}
\subsubsection{Target Declaration, State Estimation and Pruning} 
The obtained belief $\widetilde{p}(\underline{\mathbf{x}}_{k}, \underline{\mathbf{r}}_{k})$ can be used for target declaration, state estimation and pruning. Concretely, the beliefs $\widetilde{p}(\underline{\mathbf{x}}_{k}^{(i)},\underline{r}_{k}^{(i)})$ and $\widetilde{p}( \underline{r}_{k}^{(i)})$ approximating the pdf $p(\underline{\mathbf{x}}_{k}^{(i)},\underline{r}_{k}^{(i)}|\mathbf{z}_{1:k})$ and the pmf  $p(\underline{r}_{k}^{(i)}|\mathbf{z}_{1:k})$ are derived by
\begin{align*}
	\widetilde{p}(\underline{\mathbf{x}}_{k}^{(i)}, \underline{r}_{k}^{(i)})&=\sum_{\underline{\mathbf{r}}_{k}\backslash r_{k}^{(i)}}\int\widetilde{p}(\underline{\mathbf{x}}_{k}, \underline{\mathbf{r}}_{k})\mathrm{d}{(\underline{\mathbf{x}}_{k}\backslash\underline{\mathbf{x}}_{k}^{(i)})},
	\\
	\widetilde{p}( \underline{r}_{k}^{(i)})&=\sum_{\underline{\mathbf{r}}_{k}\backslash r_{k}^{(i)}}\int\widetilde{p}(\underline{\mathbf{x}}_{k}, \underline{\mathbf{r}}_{k})\mathrm{d}{\underline{\mathbf{x}}_{k}}.
\end{align*}
Then, the target declaration can be performed by comparing the existence probability with a given threshold $P_{\mathrm{e}}$, i.e., the legacy PT $\underline{\mathbf{x}}_{k}^{(i)}$ is confirmed at time $k$ if $\widetilde{p}( \underline{r}_{k}^{(i)}=1)>P_{\mathrm{e}}$. By means of the
minimum mean square error (MMSE) estimator, the state estimation for these PTs are obtained as
\begin{align}\label{mse}
	\widehat{\underline{\mathbf{x}}}_{k}^{(i)}=\int\underline{\mathbf{x}}_{k}^{(i)}\frac{\widetilde{p}(\underline{\mathbf{x}}_{k}^{(i)}, \underline{r}_{k}^{(i)})}{\widetilde{p}( \underline{r}_{k}^{(i)})}\mathrm{d}{\underline{\mathbf{x}}}_{k}^{(i)}.
\end{align} 
Analogously, the implementation of the target declaration and the state estimation for the new PTs are the same as for the legacy PTs. Finally, a pruning step is performed to remove unlikely PTs. Specifically, let $P_{\mathrm{pr}}$ be the pruning threshold, and then the PTs with existence beliefs smaller than $P_{\mathrm{pr}}$ are removed, i.e., the legacy PTs with $\widetilde{p}( \underline{r}_{k}^{(i)}=1)<P_{\mathrm{pr}}$ and the new PTs with $\widetilde{p}(\overline{r}_{k}^{(m)}=1)<P_{\mathrm{pr}}$.
\subsection{Computational Complexity and Scalability}\label{sec-scalability}
As a highly efficient and flexible algorithm, BP provides a scalable solution to the data association problem. By exploiting the scalability of BP, we propose a GTBP method for GTT. Under the assumption of a fixed number of BP iterations, we analyze the computational complexity of the proposed GTBP method as follows. Specifically, for the prediction of the group structure, (\ref{alpha-gs}) requires to be calculated $|\underline{\mathcal{G}}_{k}|$ times. That is, its computational complexity scales as $\mathcal{O}(|\underline{\mathcal{G}}_{k}|)$. In the calculation of (\ref{app-pdf}), the computational complexity is linear in the number of group partitions. Furthermore, the computational complexity of (\ref{beta})-(\ref{gamma_k}) scales as $\mathcal{O}(n_km_k)$, where the number of measurements $m_k$ increases linearly with the number of legacy PTs, new PTs and false alarms. Notably, the worst case is that the number of PTs increases up to the maximum possible number of PTs $N_{\text{max}}$. Consequently, the overall computational complexity of GTBP scales linearly in the number of group partitions and quadratically in the number of legacy PTs. 

It is worth noting that the computational complexity can be further reduced in different ways, e.g., gating preprocessing of targets and measurements \cite{Mallick2013}, censoring of messages \cite{BP-ETT1} and preserving the $M$-best group partitions, etc. Specifically, gating technology can be used to keep the number of considered group partitions $|\underline{\mathcal{G}}_{k}|$ and the size of iterative data association at a tractable level. Message censoring ignores these messages related to new PTs that are unlikely to be an actual target. Preserving the $M$-best group partitions at each time step reduces the computational complexity of calculating the messages that involve the summation over possible group partitions (e.g., (\ref{alpha-gs}), (\ref{beta}) and (\ref{app-pdf})).
\section{Particle-based GTBP Implementation}
For general nonlinear and non-Gaussian dynamic system, it is not possible to obtain an analytical expression for the integral calculation of the aforementioned messages and beliefs. In this section, we consider an approximate particle implementation of the proposed GTBP method. Assume that the belief $\widetilde{p}(\mathbf{x}_{k-1},\mathbf{r}_{k-1})$ at time $k-1$ is approximated by a set of weighted particles $\{\{(\mathbf{x}_{k-1}^{(i,l)}, w_{k-1}^{(i,l)})\}_{i=1}^{n_k}\}_{l=1}^{L}$, where $L$ is the number of particles. Note that the summarization $\sum_{l=1}^{L}w_{k-1}^{(i,l)}$ provides an approximation of the marginal posterior pmf $p(r_{k-1}^{(i)}=1|\mathbf{z}_{1:k-1})$. Specific calculations of the above messages and beliefs using particles are given as follows.
\subsection{Prediction}
For each possible group partition $\underline{\mathbf{g}}_{k}\in\underline{\mathcal{G}}_{k}$, an approximation $	\widetilde{\alpha}_k(\underline{\mathbf{g}}_{k})$ of the message $\alpha_k(\underline{\mathbf{g}}_{k})$ (\ref{pre-gs}) is calculated via the weighted particles,
\begin{align}\label{alpha_g}
	\begin{split}
		\widetilde{\alpha}_k(\underline{\mathbf{g}}_{k})&=\widetilde{C}\prod_{i\in\Lambda_{\underline{\mathbf{g}}_{k}}}\big(P_0(1-P_0)^{N(\underline{\mathbf{g}}_{k})-1}(1-\sum_{l=1}^{L}w_{k-1}^{(i,l)})+\sum_{l=1}^{L}w_{k-1}^{(i,l)}P_{i,\underline{g}_{k}^{(i)}}^{(l)}\prod_{j\in \{1,\ldots,N(\underline{\mathbf{g}}_{k})\}\backslash\underline{g}_{k}^{(i)}}(1-P_{i,j}^{(l)})\big),
	\end{split}
\end{align}
where $\widetilde{C}$ is a normalization constant such that $\sum_{\underline{\mathbf{g}}_{k}\in\underline{\mathcal{G}}_{k}}\widetilde{\alpha}_k(\underline{\mathbf{g}}_{k})=1$, and $1-\sum_{l=1}^{L}w_{k-1}^{(i,l)}$ provides an approximation
of $\int\widetilde{p}(\mathbf{x}_{k-1}^{(i)},r_{k-1}^{(i)}=0)\mathrm{d}{\mathbf{x}_{k-1}^{(i)}}$. The quantities $P_{i,\underline{g}_{k}^{(i)}}^{(l)}$ are calculated according to (\ref{Pij}) by using the particles $\mathbf{x}_{k-1}^{(i,l)}$, $i\in\Lambda_{\underline{\mathbf{g}}_{k}}$. Notably, the computation cost of the summarization in (\ref{alpha_g}) increases with the number of particles. As an alternative, one may approximately computing (\ref{alpha_g}) by using the state estimates to reduce the computation, i.e.,
\begin{align}\label{alpha_g_appro}
	\begin{split}
		\widetilde{\alpha}_k(\underline{\mathbf{g}}_{k})&\approx \widetilde{C}\prod_{i\in\Lambda_{\underline{\mathbf{g}}_{k}}}\big(P_0(1-P_0)^{N(\underline{\mathbf{g}}_{k})-1}(1-\sum_{l=1}^{L}w_{k-1}^{(i,l)})+(\sum_{l=1}^{L}w_{k-1}^{(i,l)})\widehat{P}_{i,\underline{g}_{k}^{(i)}}\prod_{j\in \{1,\ldots,N(\underline{\mathbf{g}}_{k})\}\backslash\underline{g}_{k}^{(i)}}(1-\widehat{P}_{i,j})\big),
	\end{split}
\end{align}
where $\widehat{P}_{i,j}$ are computed by using the state estimates. As described in Subsection \ref{sec-scalability}, we also can apply the $M$-best strategy to further reduce the computational complexity, i.e., preserving the $M$ most likely group partitions and renormalizing the preserved messages $\widetilde{\alpha}_k(\underline{\mathbf{g}}_{k})$. To simplify notations, we redefine $\underline{\mathcal{G}}_{k}:=\{1,\ldots,M\}$ as an index set of the preserved group partitions at time $k$, which can be easily implemented by associating each preserved $\underline{\mathbf{g}}_{k}$ with a unique index $g\in\underline{\mathcal{G}}_{k}$. By replacing  $\widetilde{p}(\mathbf{x}_{k-1},\mathbf{r}_{k-1})$ in (\ref{alpha1}) with particles, the message $\alpha_k(\underline{\mathbf{x}}_{k}, \underline{\mathbf{r}}_{k}|g)$ under given group partition $g\in\underline{\mathcal{G}}_{k}$ is approximated by
\begin{align*}
	\alpha_k(\underline{\mathbf{x}}_{k}, \underline{\mathbf{r}}_{k}|g)\approx\prod_{i=1}^{n_k}\widetilde{\alpha}_k(\underline{\mathbf{x}}_{k}^{(i)}, \underline{r}_{k}^{(i)}|g),
\end{align*}
where $\widetilde{\alpha}_k(\underline{\mathbf{x}}_{k}^{(i)}, 1|g)$  is represented by a set of weighted particles $\{\underline{\mathbf{x}}_{k}^{(i,l,g)},\underline{w}_{k,*}^{(i,l,g)}\}_{l=1}^{L}$. More specifically, for the PTs belonging to the groups $j\neq0$ in the group partition $g$, the particles $\underline{\mathbf{x}}_{k}^{(i,l,g)}$ are drawn from the group transition density (\ref{pdf-model}), where the offsets and the virtual leaders are calculated by using corresponding particles at time $k-1$ according to (\ref{offset})-(\ref{vl}). Otherwise, the particles $\underline{\mathbf{x}}_{k}^{(i,l,g)}$ are drawn from the single-target state transition density in (\ref{sg}). Furthermore, the weight $\underline{w}_{k,*}^{(i,l,g)}$ is updated by
\begin{align}\label{w1}
	\underline{w}_{k,*}^{(i,l,g)}=p_{\mathrm{s}}(\mathbf{x}_{k-1}^{(i,l)})w_{k-1}^{(i,l)}.
\end{align}
\subsection{Measurement Evaluation, Update and Belief Calculation}
Next, an approximation $\widetilde{\beta}_k(a_{k}^{(i)})$ of the message $\beta_k(a_{k}^{(i)})$ in (\ref{beta}) can be calculated from the weighted particles $\{\{\underline{\mathbf{x}}_{k}^{(i,l,g)},\underline{w}_{k,*}^{(i,l,g)}\}_{l=1}^{L}\}_{g=1}^{M}$,
\begin{align}\label{app-beta}
	\begin{split}
		\widetilde{\beta}_k(a_{k}^{(i)})&= \sum_{g=1}^{M}\widetilde{\alpha}_k(g)\sum_{l=1}^{L}q(\underline{\mathbf{x}}_{k}^{(i,\l,g)}, 1, a_{k}^{(i)}; \mathbf{z}_{k}) \underline{w}_{k,*}^{(i,l,g)}+\mathrm{I}(a_{k}^{(i)})\sum_{g=1}^{M}\widetilde{\alpha}_k(g)(1-\sum_{l=1}^{L}\underline{w}_{k,*}^{(i,l,g)}).
	\end{split}
\end{align}
According to (\ref{def-v}) and (\ref{xi}), we have $\xi_k(b_{k}^{(m)})=1$ for $b_{k}^{(m)}\neq0$, and
\begin{align*}
	\begin{split}
		\xi_k(0)&= \int\frac{\mu_{\mathrm{b}}f_{\mathrm{b}}(\overline{\mathbf{x}}_{k}^{(m)})f(\mathbf{z}_{k}^{(m)}| \overline{\mathbf{x}}_{k}^{(m)})}{\mu_{\mathrm{c}}f_{\mathrm{c}}(\mathbf{z}_{k}^{(m)})}\mathrm{d}{\overline{\mathbf{x}}_{k}^{(m)}}+1,
	\end{split}
\end{align*}
which can be approximated by the particles $\{\overline{\mathbf{x}}_{k}^{(m,l)}\}_{l=1}^{L}$ sampled from the prior distribution $f_{\mathrm{b}}(\overline{\mathbf{x}}_{k}^{(m)})$ with weights $\overline{w}_{k,*}^{(m,l)}$, i.e.,
\begin{align}\label{app-xi}
	\begin{split}
		\widetilde{\xi}_k(0)&= \frac{\mu_{\mathrm{b}}}{\mu_{\mathrm{c}}f_{\mathrm{c}}(\mathbf{z}_{k}^{(m)})}\sum_{l=1}^{L}	\overline{w}_{k,*}^{(m,l)}f(\mathbf{z}_{k}^{(m)}| \overline{\mathbf{x}}_{k}^{(m,l)})+1.
	\end{split}
\end{align}
The approximate messages $\widetilde{\beta}_k(a_{k}^{(i)})$ and $\widetilde{\xi}_k(b_{k}^{(m)})$ obtained above are substituted
for corresponding messages in the iterative data association step (\ref{message-beta})-(\ref{ini-beta}). After the iterations terminate, approximate messages $\widetilde{\kappa}_k(a_{k}^{(i)})$ and $\widetilde{\iota}_k(b_{k}^{(m)})$ of (\ref{kappa}) and (\ref{iota}) are derived. Then, the approximate messages $\widetilde{\gamma}_k^{(i)}(\underline{\mathbf{x}}_k^{(i)}, \underline{r}_k^{(i)})$ are obtained as
\begin{align}\label{app-gamma}
	\begin{split}
		\widetilde{\gamma}_k^{(i)}(\underline{\mathbf{x}}_k^{(i)}, 1)&=\sum_{a_{k}^{(i)}=0}^{m_{k}} q(\underline{\mathbf{x}}_{k}^{(i)}, 1, a_{k}^{(i)}; \mathbf{z}_{k}) \widetilde{\kappa}_k(a_{k}^{(i)}),
	\end{split}
\end{align}
with $\widetilde{\gamma}_k^{(i)}(\underline{\mathbf{x}}_k^{(i)}, 0)=\widetilde{\kappa}_k(0)$. Then, an approximation of the belief $\widetilde{p}(\underline{\mathbf{x}}_{k}, \underline{\mathbf{r}}_{k})$ is obtained as
\begin{align*}
	\widetilde{p}(\underline{\mathbf{x}}_{k}, \underline{\mathbf{r}}_{k})&\propto \sum_{g=1}^{M}\widetilde{\alpha}_k(g)\prod_{i=1}^{n_{k}}\widetilde{p}(\underline{\mathbf{x}}_{k}^{(i)}, \underline{r}_{k}^{(i)}|g),
\end{align*}
with 
\begin{align*}
	\widetilde{p}(\underline{\mathbf{x}}_{k}^{(i)}, \underline{r}_{k}^{(i)}|g):=\widetilde{\alpha}_k(\underline{\mathbf{x}}_{k}^{(i)}, \underline{r}_{k}^{(i)}|g)\widetilde{\gamma}_k^{(i)}(\underline{\mathbf{x}}_k^{(i)}, \underline{r}_k^{(i)}),
\end{align*} 
where $\widetilde{p}(\underline{\mathbf{x}}_{k}^{(i)}, 1|g)$ can be represented by a set of particles $\{\underline{\mathbf{x}}_{k}^{(i,l,g)}\}_{l=1}^{L}$ with the following nonnormalized weights
\begin{align}\label{wa1}
	\begin{split}
		\underline{w}_{k,*}^{A(i,l,g)}&=\underline{w}_{k,*}^{(i,l,g)}
		\widetilde{\gamma}_k^{(i)}(\underline{\mathbf{x}}_{k}^{(i,l,g)}, \underline{r}_k^{(i)})
		\\
		&=\underline{w}_{k,*}^{(i,l,g)}\sum_{a_{k}^{(i)}=0}^{m_{k}} q(\underline{\mathbf{x}}_{k}^{(i,l,g)}, 1, a_{k}^{(i)}; \mathbf{z}_{k}) \widetilde{\kappa}_k(a_{k}^{(i)}).
	\end{split}
\end{align}
Similarly, the nonnormalized
weights corresponding to $\widetilde{p}(\underline{\mathbf{x}}_{k}^{(i)}, 0|g)$ are obtained as
\begin{align}\label{wb1}
	\underline{w}_{k,*}^{B(i,g)}=(1-\sum_{l=1}^{L}\underline{w}_{k,*}^{(i,l,g)}) \widetilde{\kappa}_k(0),
\end{align}
where $1-\sum_{l=1}^{L}\underline{w}_{k,*}^{(i,l,g)}$ provides an approximation of $\int\alpha_k(\underline{\mathbf{x}}_{k}^{(i)}, 0|g)\mathrm{d}{\underline{\mathbf{x}}_{k}^{(i)}}$. Hence, $\widetilde{p}(\underline{\mathbf{x}}_{k}^{(i)}, 1|g)$ can be represented by a set of weighted particles $\{\underline{\mathbf{x}}_{k}^{(i,l,g)},\underline{w}_{k}^{(i,l,g)}\}_{l=1}^{L}$, where
\begin{align}\label{w1-update}
	\underline{w}_{k}^{(i,l,g)}=\frac{\underline{w}_{k,*}^{A(i,l,g)}}{\sum_{l=1}^{L}\underline{w}_{k,*}^{A(i,l,g)}+\underline{w}_{k,*}^{B(i,g)}}.
\end{align}
Thus, the particle-based approximation of
the posterior pmfs $p(\underline{r}_{k}^{(i)}=1|\mathbf{z}_{1:k})$ are obtained as 
\begin{align}\label{ex-legacy}
	\widetilde{p}( \underline{r}_{k}^{(i)})\approx\sum_{g=1}^{M}\widetilde{\alpha}_k(g)\sum_{l=1}^{L}\underline{w}_{k}^{(i,l,g)}.
\end{align}
According to (\ref{mse}), the state estimation for the legacy PTs $\underline{\mathbf{x}}_{k}^{(i)}$ are obtained as
\begin{align}\label{est-legacy}
	\widehat{\underline{\mathbf{x}}}_{k}^{(i)}=\sum_{g=1}^{M}\sum_{l=1}^{L}\frac{\widetilde{\alpha}_k(g)\underline{w}_{k}^{(i,l,g)}\underline{\mathbf{x}}_{k}^{(i,l,g)}}{\sum_{g=1}^{M}\widetilde{\alpha}_k(g)\sum_{l=1}^{L}\underline{w}_{k}^{(i,l,g)}}.
\end{align} 
Furthermore, the beliefs $\widetilde{p}(g)$ approximating the marginal posterior pmfs $p(g|\mathbf{z}_{1:k})$ in (\ref{pdf-g}) are obtained as
\begin{align*}
	\widetilde{p}(g)=\frac{\widetilde{\alpha}_k(g)\prod\limits_{i=1}^{n_{k}}\sum\limits_{\underline{r}_{k}^{(i)}\in\{0,1\}}\int\widetilde{p}(\underline{\mathbf{x}}_{k}^{(i)},\underline{r}_{k}^{(i)}|g)\mathrm{d}{\underline{\mathbf{x}}_{k}^{(i)}}}{\sum\limits_{g=1}^{M}\widetilde{\alpha}_k(g)\prod\limits_{i=1}^{n_{k}}\sum\limits_{\underline{r}_{k}^{(i)}\in\{0,1\}}\int\widetilde{p}(\underline{\mathbf{x}}_{k}^{(i)},\underline{r}_{k}^{(i)}|g)\mathrm{d}{\underline{\mathbf{x}}_{k}^{(i)}}},
\end{align*}
where the particle-based approximation are given by
\begin{align}\label{est-g}
	\widetilde{p}(g)=\frac{\widetilde{\alpha}_k(g)\prod\limits_{i=1}^{n_{k}}(\sum_{l=1}^{L}\underline{w}_{k,*}^{A(i,l,g)}+\underline{w}_{k,*}^{B(i,g)})}{\sum\limits_{g=1}^{M}\widetilde{\alpha}_k(g)\prod\limits_{i=1}^{n_{k}}(\sum_{l=1}^{L}\underline{w}_{k,*}^{A(i,l,g)}+\underline{w}_{k,*}^{B(i,g)})}.
\end{align}
Note that for each PT $i$ at time $k-1$, $L$ particles $\{\mathbf{x}_{k-1}^{(i,l)}\}_{l=1}^{L}$ are propagated to $L\times M$ particles $\{\{\underline{\mathbf{x}}_{k}^{(i,l,g)}\}_{l=1}^{L}\}_{g=1}^{M}$ based on the $M$-best group partitions. To reduce the $L\times M$ particles to $L$ particles, a resampling step is performed according to the beliefs $\widetilde{p}(g)$.

For the new PTs, a particle-based approximation $\widetilde{\varsigma}_k^{(m)}(\overline{\mathbf{x}}_k^{(m)}, 1)$ of the messages $\varsigma_k^{(m)}(\overline{\mathbf{x}}_k^{(m)}, 1)$ in (\ref{varsigma}) is given by $\{\overline{\mathbf{x}}_{k}^{(m,l)}\}_{l=1}^{L}$ with nonnormalized weights
\begin{align}\label{wa2}
	\overline{w}_{k,*}^{A(m,l)}=\overline{w}_{k,*}^{(m,l)}\times\frac{\mu_{\mathrm{b}}f(\mathbf{z}_{k}^{(m)}| \overline{\mathbf{x}}_{k}^{(m,l)})\widetilde{\iota}_k(0)}{\mu_{\mathrm{c}}f_{\mathrm{c}}(\mathbf{z}_{k}^{(m)})},
\end{align}
and the nonnormalized weights corresponding to $\widetilde{\varsigma}_k^{(m)}(\overline{\mathbf{x}}_k^{(m)}, 0)$ are given by
\begin{align}\label{wb2}
	\overline{w}_{k,*}^{B(m,l)}=\sum_{b_{k}^{(m)}=0}^{n_k}\widetilde{\iota}_k(b_k^{(m)}).
\end{align}
Thus, the beliefs $\widetilde{p}(\overline{\mathbf{x}}_{k}^{(m)}, 1)$ in (\ref{new-pt}) are approximated by the particles $\{(\overline{\mathbf{x}}_{k}^{(m,l)}, \overline{w}_{k}^{(m,l)})\}_{l=1}^{L}$, where
\begin{align}\label{w2-update}
	\overline{w}_{k}^{(m,l)}&=\frac{\overline{w}_{k,*}^{A(m,l)}}{\sum_{l=1}^{L}\overline{w}_{k,*}^{A(m,l)}+\overline{w}_{k,*}^{B(m,l)}}.
\end{align}
Then, the marginal posterior pmfs $p(\overline{r}_{k}^{(m)}=1|\mathbf{z}_{1:k})$ are approximated by
\begin{align}\label{ex-new}
	\widetilde{p}( \overline{r}_{k}^{(m)})\approx\sum_{l=1}^{L}\underline{w}_{k}^{(m,l)},
\end{align}
and the state estimation for the new PT $\overline{\mathbf{x}}_{k}^{(m)}$ are obtained as
\begin{align}\label{est-new}
	\widehat{\overline{\mathbf{x}}}_{k}^{(m)}=\sum_{l=1}^{L}\frac{\overline{w}_{k}^{(m,l)}\overline{\mathbf{x}}_{k}^{(m,l)}}{\sum_{l=1}^{L}\overline{w}_{k}^{(m,l)}}.
\end{align}
A pseudo-code description
of the particle-based implementation of $M$-best GTBP is summarized as follows, which preserves the $M$-best group partitions at each time step. For notational convenience, we ignore the changes in the indices of the legacy PT $i$ and the new PT $m$ before and after pruning in Algorithm \ref{A1}.
\begin{algorithm}[ht]
	\caption{Particle-based Implementation of the $M$-best GTBP Algorithm}
	\label{A1}
	\begin{algorithmic}[0]
		\renewcommand{\algorithmicrequire}{\textbf{Initialize:}}
		\Require \\
		Set $\mathbf{y}_0$ and $\mathbf{g}_0$ as empty vectors;
		\renewcommand{\algorithmicrequire}{\textbf{Input at time $k$:}}
		\Require \\
		Weighted particles $\{\{(\mathbf{x}_{k-1}^{(i,l)}, w_{k-1}^{(i,l)})\}_{i=1}^{n_k}\}_{l=1}^{L}$, and measurements $\mathbf{z}_{k}$;
		\renewcommand{\algorithmicrequire}{\textbf{Output at time $k$:}}
		\Require \\
		Legacy PTs : state estimation  $\widehat{\underline{\mathbf{x}}}_{k}^{(i)}$, beliefs $\widetilde{p}( \underline{r}_{k}^{(i)})$ and weighted particles $\{\{(\underline{\mathbf{x}}_{k}^{(i,l)}, \underline{w}_{k}^{(i,l)})\}_{i=1}^{n_{k}}\}_{l=1}^{L}$;\\
		New PTs: state estimation  $\widehat{\overline{\mathbf{x}}}_{k}^{(m)}$, beliefs $\widetilde{p}( \overline{r}_{k}^{(m)})$ and weighted particles $\{\{(\overline{\mathbf{x}}_{k}^{(m,l)},\overline{w}_{k}^{(m,l)})\}_{m=1}^{m_k}\}_{l=1}^{L}$;\\
		Group structure: the preserved $M$-best group structures and corresponding probabilities $\widetilde{p}(g)$;
		\renewcommand{\algorithmicrequire}{\textbf{Run:}}
		\Require 
		\State {\bf Step 1:} compute $\widetilde{\alpha}_k(g)$, $g\in\underline{\mathcal{G}}_k$ via (\ref{alpha_g_appro}), preserve the $M$ most likely group partitions and renormalize $\widetilde{\alpha}_k(g)$;
		
		\State {\bf Step 2:} for each group partition $g$, draw the particles $\underline{\mathbf{x}}_{k}^{(i,l,g)}, l=1,\ldots,L$ from the group transition density $(\ref{pdf-model})$ or the single-target state transition density in (\ref{sg}), and compute corresponding weights $\underline{w}_{k,*}^{(i,l,g)}$ via (\ref{w1});
		
		\State {\bf Step 3:} compute $\widetilde{\beta}_k(a_{k}^{(i)})$, $a_{k}^{(i)}=0,\ldots,m_k$ via (\ref{app-beta}), draw particles with equal weights from the prior pdf $f_{\mathrm{b}}(\overline{\mathbf{x}}_{k}^{(m)})$, i.e., $\{(\overline{\mathbf{x}}_{k}^{(m,l)}, \overline{w}_{k,*}^{(m,l)}=\frac{1}{L})\}_{l=1}^{L}$, and compute $\widetilde{\xi}_k(b_{k}^{(m)}=0)$ via (\ref{app-xi});
		
		\State {\bf Step 4:} run the iterative data association (\ref{message-beta})-(\ref{ini-beta}), and compute $\widetilde{\kappa}_k(a_{k}^{(i)})$ and $\widetilde{\iota}_k(b_{k}^{(m)})$ via (\ref{kappa})-(\ref{iota});
		
		\State {\bf Step 5:} for the legacy PTs $i\in\{1,\ldots,n_k\}$, calculate the weights $\underline{w}_{k}^{(i,l,g)}$ via (\ref{wa1})-(\ref{w1-update}), and then obtain $\widetilde{p}( \underline{r}_{k}^{(i)})$, $\widehat{\underline{\mathbf{x}}}_{k}^{(i)}$, $\widetilde{p}(g)$ via (\ref{ex-legacy})-(\ref{est-g}), respectively;
		
		\State {\bf Step 6:} for the new PTs $m\in\{1,\ldots,m_k\}$, calculate the weights $\overline{w}_{k}^{(m,l)}$ via (\ref{wa2})-(\ref{w2-update}), and then obtain $\widetilde{p}( \overline{r}_{k}^{(m)})$, $\widehat{\overline{\mathbf{x}}}_{k}^{(m)}$ via (\ref{ex-new}) and (\ref{est-new}), respectively;
		
		\State {\bf Step 7:} prune the legacy PTs and new PTs with existence probabilities less than the threshold $P_{\mathrm{pr}}$;
		
		\State {\bf Step 8:} according to the probability $\widetilde{p}(g)$, a resample step for each preserved legacy PT is performed to reduce the $L\times M$ particles $\{\{\underline{\mathbf{x}}_{k}^{(i,l,g)}\}_{g=1}^{M}\}_{l=1}^{L}$ to $L$ particles  $\{\underline{\mathbf{x}}_{k}^{(i,l)}\}_{l=1}^{L}$ with equal wights $ \underline{w}_{k}^{(i,l)}=\frac{\widetilde{p}(\underline{r}_{k}^{(i)})}{L}$;
		
		\\
		\Return $\{\{(\underline{\mathbf{x}}_{k}^{(i,l)}, \underline{w}_{k}^{(i,l)})\}_{l=1}^{L},\ \widetilde{p}( \underline{r}_{k}^{(i)}),\  \widehat{\underline{\mathbf{x}}}_{k}^{(i)}\}_{i=1}^{n_{k}}$,  $\widetilde{p}(g)$, $\{(\overline{\mathbf{x}}_{k}^{(m,l)},\overline{w}_{k}^{(m,l)})\}_{l=1}^{L},\ \widetilde{p}( \overline{r}_{k}^{(m)}),\  \widehat{\overline{\mathbf{x}}}_{k}^{(m)}\}_{m=1}^{m_k}$;
	\end{algorithmic}
\end{algorithm}
\section{Simulation}
In this section, we simulate two typical GTT scenarios to demonstrate
the performance of the proposed GTBP method. The simulation setting and performance comparison results are presented as follows.
\subsection{Simulation Setting}
In scenario 1, we consider tracking an unknown number of group targets, involving the group splitting and merging. A total of 100 time steps with the time sampling interval $\Delta T = 2s$ is simulated, and four targets appear in the scene. Let the kinematics of the individual targets described by the state vector $\mathbf{x}_k^{(i)}=\left[x_k^{(i)},\dot{x}_k^{(i)},y_k^{(i)},\dot{y}_k^{(i)}\right]^{\mathrm{T}}$ of planar position and velocity. Here, we use one constant velocity (CV) model and two constant turn (CT) models \cite{Mallick2013} without process noise to generate the true trajectories of the four targets, where the state transition matrices of the corresponding models are
\begin{align*}
	F_{\mathrm{CV}}:=\left[\begin{array}{cccc}
		1 & \Delta T & 0 & 0 \\
		0 & 1 & 0 & 0 \\
		0 & 0 & 1 & \Delta T \\
		0 & 0 & 0 & 1
	\end{array}\right],\quad F_{\mathrm{CT}}^{(j)}:=\left[\begin{array}{cccc}
	1 & \frac{\sin \omega^{(j)} \Delta T}{\omega^{(j)}} & 0 & -\frac{1-\cos \omega^{(j)} \Delta T}{\omega^{(j)}} \\
	0 & \cos \omega^{(j)} \Delta T & 0 & -\sin \omega^{(j)} \Delta T \\
	0 & \frac{1-\cos \omega^{(j)} \Delta T}{\omega^{(j)}} & 1 & \frac{\sin \omega^{(j)} \Delta T}{\omega^{(j)}} \\
	0 & \sin \omega^{(j)} \Delta T & 0 & \cos \omega^{(j)} \Delta T
\end{array}\right],
\end{align*}
respectively, where $j\in\{1,2\}$, the turn rates $\omega^{(1)}=2.25^{\circ}/s$ and $\omega^{(2)}=-2.25^{\circ}/s$.

\begin{figure}[H]
	\centering 
	\includegraphics[width=0.6\linewidth]{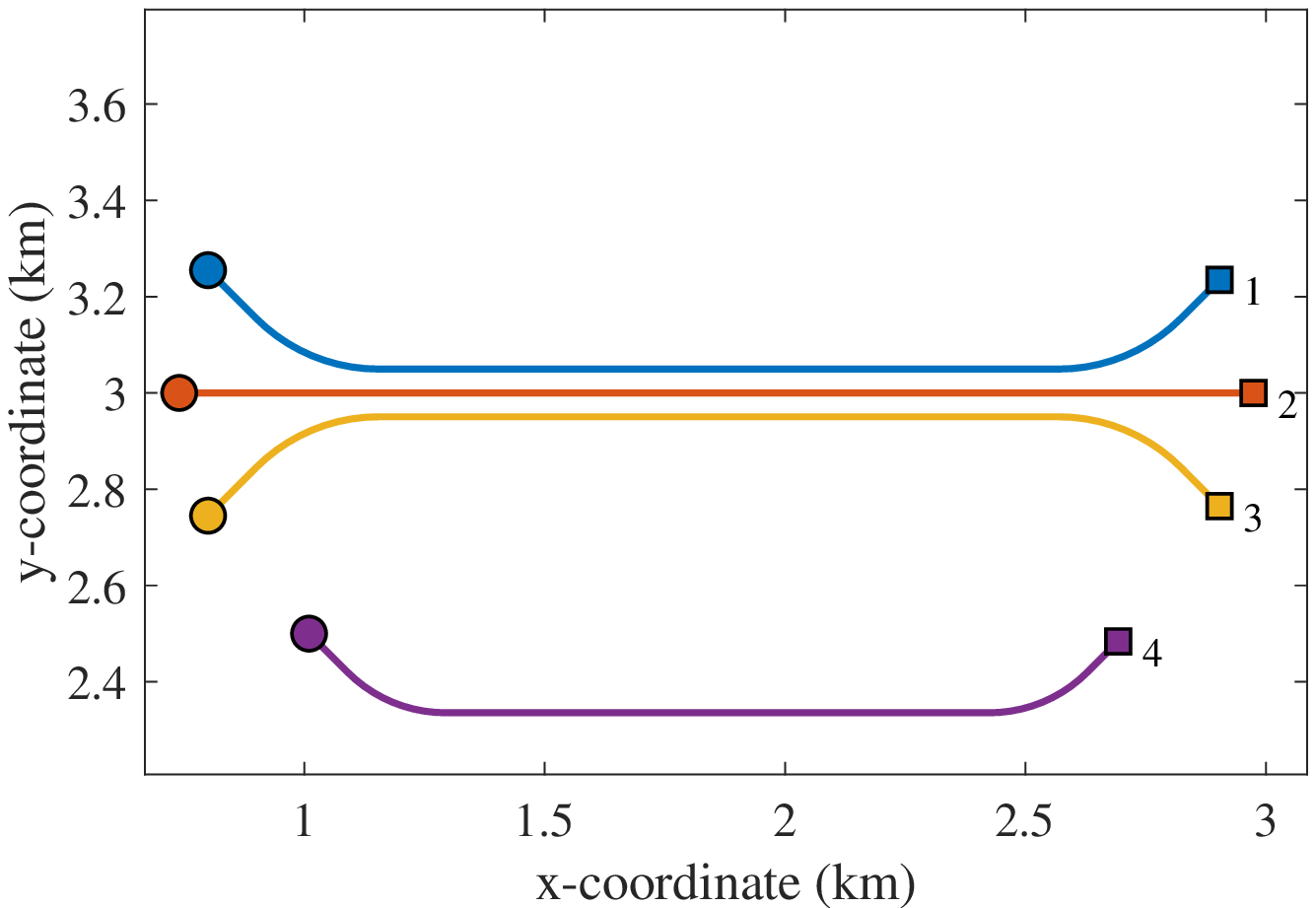}
	\caption{The ground truths simulated in scenario 1. Starting and stopping positions are marked with $\circ$ and $\square$, respectively.}
	\label{figure3}
\end{figure}
\begin{table}[H]
	\centering
	\caption{The lifespan (time step) and initial speeds of the targets.}
	\label{table1} 
	\begin{tabular}{ccc} 
		\toprule 
		Target indices & Lifespan & Initial states
		\\ 
		\midrule 
		1 &$\left[1,80\right]$ & $\left[800, 10, 3255, -10\right]^{\mathrm{T}}$
		\\
		2 &$\left[1,80\right]$ &$\left[740, 10\sqrt{2}, 3000, 0\right]^{\mathrm{T}}$
		\\
		3 &$\left[1,80\right]$ &$\left[800, 10, 2745, 10\right]^{\mathrm{T}}$
		\\
		4 &$\left[21,100\right]$ &$\left[1010, 8, 2500, -8\right]^{\mathrm{T}}$
		\\
		\bottomrule
	\end{tabular} 
\end{table}
The simulated ground truths, lifespan and initial state vectors are shown in Fig. \ref{figure3} and Table \ref{table1}, respectively. As shown in Fig. \ref{figure3}, the targets 1, 2 and 3 fly towards the directions of their velocities based on the CV model, before executing the coordinated turn motions based on the CT model, which makes the three targets get closer and then merge into one group. Then, the group target move in a triangular formation based on the CV model, accompanied by the birth of the target 4. Followed by the splitting of the group, the targets 1, 2 and 3 gradually move away from each other. Finally, the simulated scenario ends with the termination of the target 4.

Assuming that the sensor is located at the origin, and the ranges of radius and azimuth are 0-5000$m$ and 0-$2\pi$ $rad$, respectively. The measurement likelihood is given by $f(\mathbf{z}_{k}^{(m)}|\underline{\mathbf{x}}_{k}^{(i)})=\mathcal{N}(\mathbf{z}_{k}^{(m)};H\mathbf{x}_{k}^{(i)},\sigma_{\mathbf{w}}^2I_2)$, where
\begin{align*}
	H:=\left[\begin{array}{cccc}
		1 & 0 & 0 & 0 \\
		0 & 0 & 1 & 0
	\end{array}\right],\quad I_2:=\left[\begin{array}{cc}
		1 & 0 \\
		0 & 1
	\end{array}\right],
\end{align*}
and the standard deviation (Std) of the measurement noise is $\sigma_{\mathbf{w}}=10m$. The clutter pdf $f_{\mathrm{c}}(\mathbf{z}_{k}^{(m)})$ is assumed uniform on the surveillance region, and the Poisson mean number of clutters is $\mu_{\mathrm{c}}=10$ if not noted otherwise.
\subsection{Simulated Methods and Performance Metric}
We compare the recently developed BP method \cite{BP-MTT2} and the proposed GTBP method. Moreover, the performance of GTBP preserving different numbers of group partitions are also tested. For notational convenience, we abbreviate the method implemented by Algorithm \ref{A1} and $M=2$ as GTBP-2best. In order to evaluate the tracking algorithm exclusively, we employ the same birth pdf $f_{\mathrm{b}}(\overline{\mathbf{x}}_{k}^{(m)})$ for all tested methods, which is constructed by using the measurements at the previous time step \cite{birth-pdf}. If not noted otherwise, we set the Poisson mean number of new PTs, the maximum possible number of PTs, the number of particles (for representing each legacy PT or new PT state), the detection probability and survival probability as $\mu_{\mathrm{b}}=10^{-5}\times\mu_{\mathrm{c}}$, $N_{\text{max}} = 8$, $L = 3000$,  $p_{\mathrm{d}}(\underline{\mathbf{x}}_{k}^{(i)})=0.995$ and $p_{\mathrm{s}}(\underline{\mathbf{x}}_{k}^{(i)})=0.9999$, respectively. The grouping constant $P_{0}$ in (\ref{Pij}) for incorporating nonexistent PTs into groups is set to $0.001$. In the iterative data association, the iteration is stopped if the Frobenius norm of the beliefs between two consecutive iterations is less than $10^{-5}$ or reaching the maximum number of iterations 100. All tested methods perform a message censoring step \cite{BP-ETT1} with a threshold 0.9. The thresholds for target declaration and pruning are $P_{\mathrm{e}}=0.8$ and $P_{\mathrm{pr}}=10^{-5}\times\mu_{\mathrm{c}}$, respectively. Furthermore, we use the CV model with the process noise $Q_{k}=\sigma_{\mathbf{v}}^2GG^{\mathrm{T}}$ for tracking, where $\sigma_{\mathbf{v}}=10m/s^2$ is the Std of the process noise and
\begin{align*}
	G:=\left[\begin{array}{cccc}
		\frac{\Delta T^{2}}{2} & \Delta T &0 &0\\
		0 & 0 & \frac{\Delta T^{2}}{2} &\Delta T
	\end{array}\right]^{\mathrm{T}}.
\end{align*}


To evaluate the tracking performance, we use the OSPA$^{(2)}$ distance $\check{d}_{p, q}^{(c)}(X, Y ; w)$ as the performance metric, which is able to capture different kinds of tracking errors such as track switching and fragmentation \cite{ospa2}. If not noted otherwise, the cutoff parameter, the order parameters and the window length are set to $c=50$, $p=1$, $q=2$ and $w=10$ (with uniform weights), respectively.
\subsection{Simulation Results of Scenario 1}
\begin{figure}[ht]
	\centering 
	\includegraphics[width=0.6\linewidth]{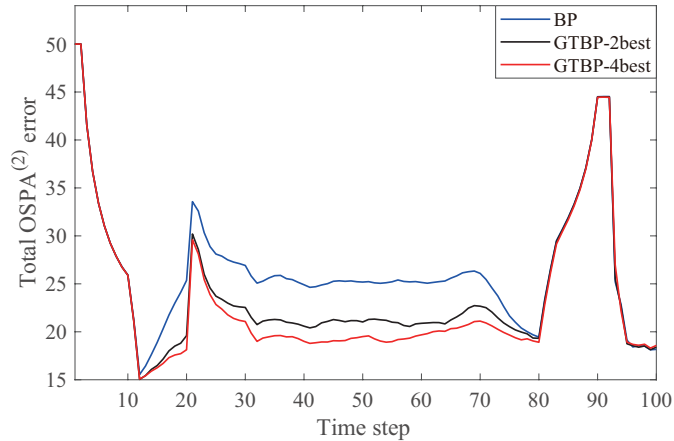}
	\caption{The average total OSPA$^{(2)}$.}
	\label{figure4}
\end{figure}

Fig. \ref{figure4} plots the average total OSPA$^{(2)}$ of BP, GTBP-2best and GTBP-4best over 100 Monte Carlo runs versus the time step. Specific results and reasons are given as follows:
\begin{itemize}
	\item Before the time step 10, it shows that the three methods have similar	performance for the reason of using the same track initialization settings and the targets 1, 2 and 3 move as ungrouped targets.
	
	\item Between the time steps 10 and 80 (i.e., the GTT stage with the occurrence of group merging and splitting), GTBP-2best and GTBP-4best outperform BP for the reason of estimating the uncertainty of the group structure. In addition, GTBP-4best outperforms GTBP-2best as a result of preserving more group partitions at each time step.
	
	\item After the time step 80, the average total OSPA$^{(2)}$ of the three methods gradually become the same, since there is only the ungrouped target 4 in the scene and thus the proposed GTBP method degenerates to the BP method.
\end{itemize}
Furthermore, the two spikes around the time steps $k=20$ and $k=80$ are caused by the windowing effects, the track initiation and termination delays.

\begin{figure}[H]
	\centering 
	\includegraphics[width=0.6\linewidth]{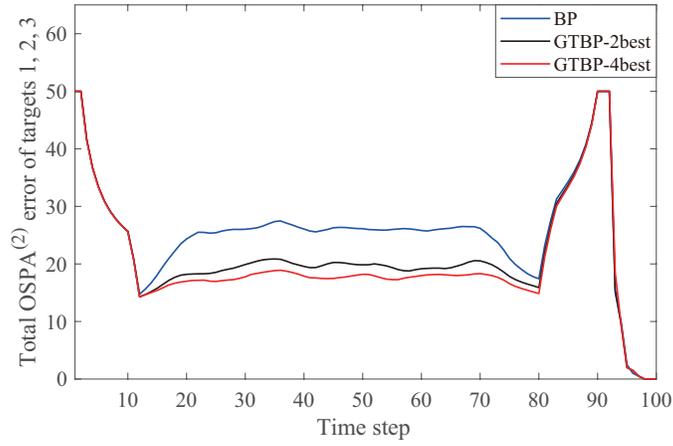}
	\caption{The average OSPA$^{(2)}$ of the group target, including the targets 1, 2, 3.}
	\label{figure5}
\end{figure}

\begin{figure}[H]
	\centering 
	\includegraphics[width=0.6\linewidth]{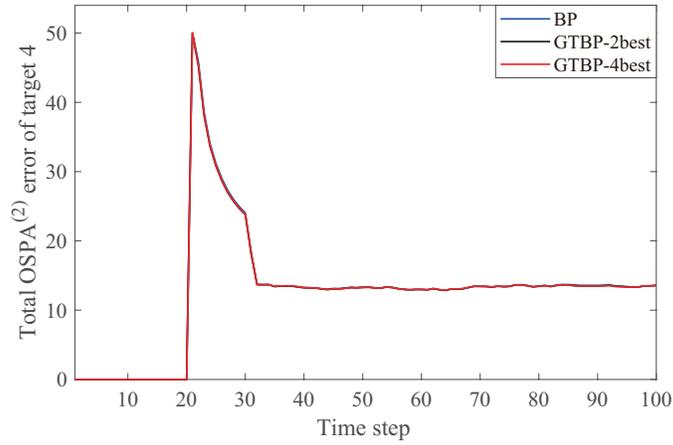}
	\caption{The average OSPA$^{(2)}$ of the  target 4.}
	\label{figure6}
\end{figure}

In more detail, we plot the average OSPA$^{(2)}$ for the group target (including the targets 1, 2 and 3) and the ungrouped target 4 in Figs. \ref{figure5}-\ref{figure6}, respectively. Fig. \ref{figure5} shows that GTBP-2best and GTBP-4best outperform BP when tracking the group target. The reasons are the same as that for the results in Fig. \ref{figure4}. Furthermore, Fig. \ref{figure6} shows that the three methods have almost the same performance when tracking the ungrouped target 4, which validates the fact that GTBP degrades to the classical BP method in the case of tracking ungrouped targets.
\subsection{Simulation Results of Scenario 2}
\begin{figure}[hbtp]
	\centering 
	\includegraphics[width=0.6\linewidth]{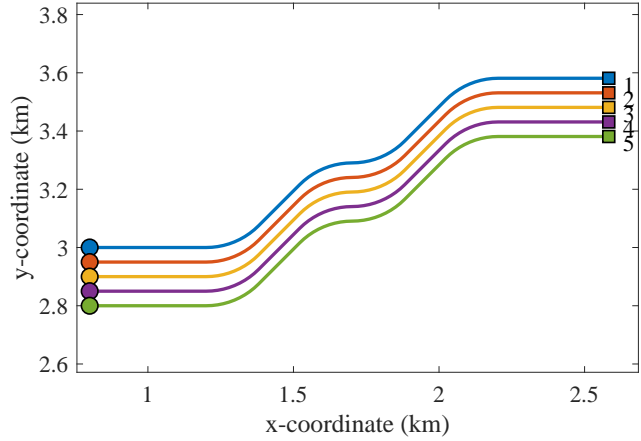}
	\caption{The ground truths simulated in scenario 2. Starting and stopping positions are marked with $\circ$ and $\square$, respectively.}
	\label{figure7}
\end{figure}

To further evaluate the performance of the proposed GTBP method, we simulate a coordinated GTT scenario and compare the average total OSPA$^{(2)}$ and the average runtimes in different cases. In this scenario, we perform a fixed number of 20 BP iterations and use $L = 1000$ particles for all test methods. If not noted otherwise, the group consists of five targets generated by the CV model and CT models with an initial speed of 10m$/$s, where the ground truths are shown in Fig. \ref{figure7}. The initial position of the target 1 is fixed at 800 m and 3000 m along the $x$-axis and $y$-axis, respectively, and the other adjacent targets within the group are separated by 50 m along the $y$-axis. 

\begin{figure}[H]
	\centering 
	\includegraphics[width=0.6\linewidth]{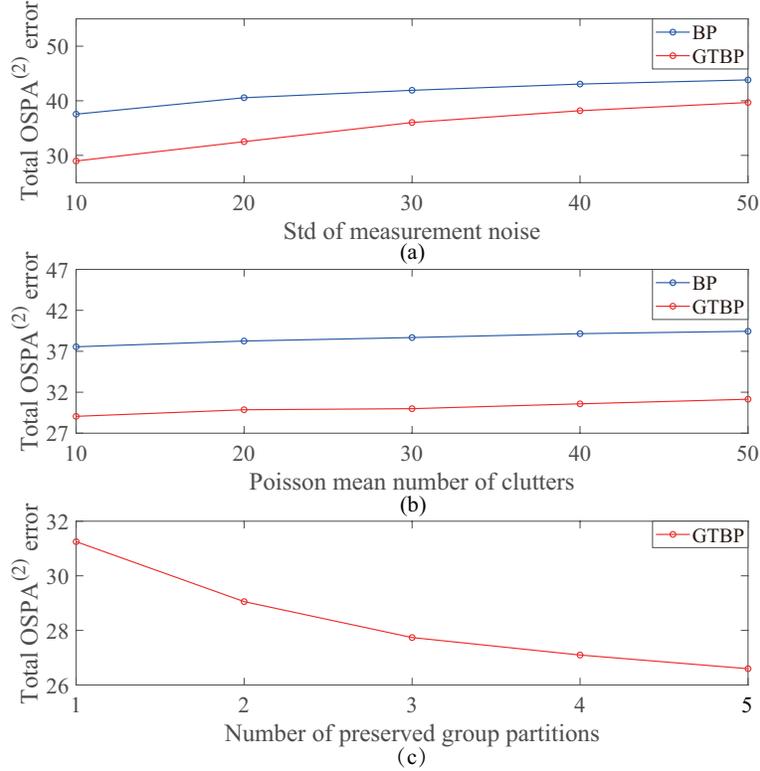}
	\caption{The average total OSPA$^{(2)}$ over 100 time steps and 100 Monte Carlo runs. (a): versus $\sigma_{\mathbf{w}}$ for 5 actual targets, $\mu_{\mathrm{c}}=10$ and $M=2$; (b): versus $\mu_{\mathrm{c}}$ for 5 actual targets, $\sigma_{\mathbf{w}}=10$ and $M=2$; (c): versus $M$ for 5 actual targets, $\sigma_{\mathbf{w}}=10$ and $\mu_{\mathrm{c}}=10$.}
	\label{figure8}
\end{figure}

Figs. \ref{figure8}a-\ref{figure8}c plot the average total OSPA$^{(2)}$ (using window length $w=20$ with uniform weights) over 100 time steps and 100 Monte Carlo runs versus the Std of the measurement noise $\sigma_{\mathbf{w}}$, the Poisson mean number of clutters $\mu_{\mathrm{c}}$ and the number of preserved group partitions $M$, respectively. More specifically,
\begin{itemize}
	\item Fig. \ref{figure8}a shows the average total OSPA$^{(2)}$ of BP and GTBP versus $\sigma_{\mathbf{w}}$ for $\mu_{\mathrm{c}}=10$ and $M=2$, which increase with the adding of $\sigma_{\mathbf{w}}$, since the target spacing is fixed and the group becomes more and more indistinguishable when increasing $\sigma_{\mathbf{w}}$. Furthermore, GTBP outperforms BP for the reason of jointly inferring the group structure uncertainty.
	
	\item Fig. \ref{figure8}b shows the average total OSPA$^{(2)}$ of BP and GTBP versus $\mu_{\mathrm{c}}$ for $\sigma_{\mathbf{w}}=10$ and $M=2$, which increase slightly as $\mu_{\mathrm{c}}$ increases. The reason may be that the number of false tracks initialized by clutters increases. Furthermore, GTBP obtains better tracking performance than BP for the same reasons analyzed in Fig. \ref{figure8}a.
	
	\item Fig. \ref{figure8}c shows the average total OSPA$^{(2)}$ of GTBP versus $M$ for $\sigma_{\mathbf{w}}=10$ and $\mu_{\mathrm{c}}=10$. The average total OSPA$^{(2)}$ decreases as the number of preserved group partitions $M$ increases, since preserving more group partitions results in a more accurate approximation of the joint posterior pdf and thus leads to further performance improvement.
\end{itemize}

\begin{figure}[H]
	\centering 
	\includegraphics[width=0.6\linewidth]{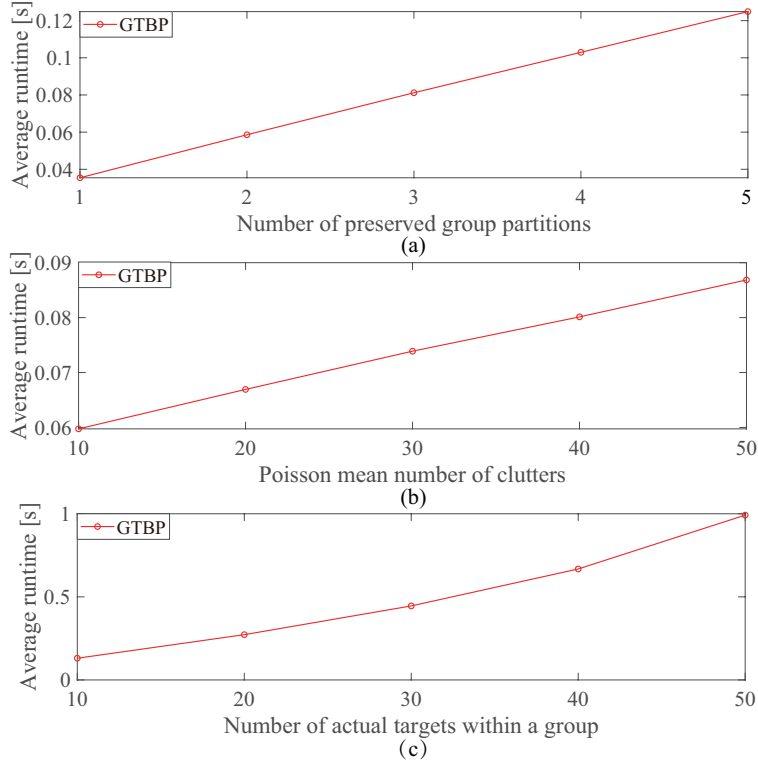}
	\caption{The average runtimes per time step of GTBP. (a): versus $M$ for 5 actual targets and $\mu_{\mathrm{c}}=10$; (b): versus $\mu_{\mathrm{c}}$ for 5 actual targets and $M=2$; (c): versus the number of actual targets for $M=2$, $\mu_{\mathrm{c}}=10$ and the maximum possible number of PTs set to $N_{\text{max}}=20, 30, \cdots, 60$.}
	\label{figure9}
\end{figure}

Furthermore, to demonstrate the excellent scalability and low complexity of the proposed GTBP method, we investigate how the runtime of GTBP scales in the number of preserved group partitions $M$, the Poisson mean number of clutters $\mu_{\mathrm{c}}$ and the number of actual targets within a group. The simulation is run on a laptop with an Intel(R) Core(TM) i5-10300H 2.50 GHz platform with 8 GB of RAM. Figs. \ref{figure9}a-\ref{figure9}c plot the average runtimes over 100 time steps and 100 Monte Carlo runs versus $M$, $\mu_{\mathrm{c}}$ and the number of actual targets within a group, respectively. The results indicate that the average runtime scales linearly in the number of preserved group partitions, linearly in the number of sensor measurements (which grows linearly with $\mu_{\mathrm{c}}$), and quadratically in the number of actual targets. Thus, GTBP has excellent scalability for GTT. Notably, the average runtime of GTBP is less than 0.1s for 5 actual targets, $M=2$ and $\mu_{\mathrm{c}}=50$, and is nearly 1s for 50 actual targets, $M=2$ and $\mu_{\mathrm{c}}=10$, which confirm that GTBP has a low complexity and it is applicable for the tracking scenarios that include large numbers of clutters and group targets.
\section{Conclusion}
In this paper, we focus on the GTT problem, where the targets within groups are closely spaced, and the groups may split and merge. We proposed a scalable GTBP method within the BP framework, which jointly infers target existence variables, group structure, data association and target states. By considering the group structure uncertainty, GTBP can capture the group structure changes such as group splitting and merging. Moreover, the introduction of group structure variables enables seamless and simultaneous tracking of multiple group targets and ungrouped targets. Specifically, the evolution of targets is modeled as the co-action of the group or single-target motions under different group structures. In particular, GTBP has excellent scalability and low complexity that only scales linearly in the numbers of preserved group partitions and sensor measurements, and quadratically in the number of targets. Numerical results verify that GTBP obtains better tracking performance and has excellent scalability in GTT. Future research direction may include the generalization of GTBP to multisensor fusion.
\bibliographystyle{ieeetr}
\bibliography{reference-BP}
\end{document}